\documentclass[]{fairmeta}
\usepackage{microtype}
\usepackage{amsfonts}
\usepackage{graphicx}
\usepackage{booktabs}
\usepackage{wrapfig}
\usepackage{amsmath}
\usepackage{amsthm}
\usepackage{algpseudocode}
\usepackage[linesnumbered,lined,boxed,commentsnumbered,ruled,longend]{algorithm2e}
\usepackage[capitalize,noabbrev]{cleveref}
\theoremstyle{plain}

\theoremstyle{definition}

\theoremstyle{remark}

\usepackage{enumitem}
\usepackage[subtle, mathdisplays=tight, charwidths=tight, leading=normal]{savetrees}
\usepackage{subcaption} 
\usepackage{booktabs} 
\usepackage{longtable} 
\usepackage{textcomp} 
\newcommand{\sysb}{\textsc{GSM}-{\large{$\infty$}}\xspace} 

\usepackage{tikz} 

\usepackage{tikz} 

\newcommand{\GSMmouse}{%
\begin{tikzpicture}[scale=0.25, line join=round, thick]
    \draw (0,0) ellipse (1.5 and 1.2);
    
    \draw (-1.2,1.1) circle (0.6);
    \draw (1.2,1.1) circle (0.6);
    
    \draw[very thick, color=pink!60] (-1.2,1.1) circle (0.4);
    \draw[very thick, color=pink!60] (1.2,1.1) circle (0.4);
    
    \fill[white] (-0.5,0.3) circle (0.25);
    \fill[white] (0.5,0.3) circle (0.25);
    \fill[black] (-0.5,0.3) circle (0.1);
    \fill[black] (0.5,0.3) circle (0.1);
    \fill[white] (-0.52,0.33) circle (0.03);
    \fill[white] (0.48,0.33) circle (0.03);
    
    \draw[line width=1.2pt, red!90] (-0.2,-0.1) circle (0.2);
    \draw[line width=1.2pt, red!90] (0.2,-0.1) circle (0.2);

    \draw (-0.1,-0.1) -- (-1.0,-0.3);
    \draw (-0.1,-0.2) -- (-1.0,-0.5);
    \draw (0.1,-0.1) -- (1.0,-0.3);
    \draw (0.1,-0.2) -- (1.0,-0.5); 
    
    \draw[line width=1.2pt, red!60] (-0.6,-0.2) .. controls (-0.2, -0.6) and (0.2, -0.6) .. (0.6,-0.2);

    \fill[red!40] (0,-0.4) ellipse (0.2 and 0.1);

\end{tikzpicture}%
}

\title{\GSMmouse\ 
{\fontsize{16pt}{10pt}\selectfont 
    GSM-{\textnormal{\scalebox{0.8}{\Huge$\infty$}}}: How Do Your LLMs Behave over
    Infinitely Increasing Context Length and Reasoning Complexity? 
} 
}

\definecolor{skyblue}{RGB}{135, 206, 235} 
\definecolor{palegreen}{RGB}{152, 251, 152}

\author{Yang Zhou$^{1*}$, Hongyi Liu$^{1*}\dag$, Zhuoming Chen$^{1}$, Yuandong Tian$^{2}$, Beidi Chen$^{1}$\\
$^1$Carnegie Mellon Univeristy\\
$^2$Meta AI\\
\texttt{\{yangzho6, zhuominc, beidic\}@andrew.cmu.edu} \\
\texttt{liuhongy21@gmail.com},
\texttt{yuandong@meta.com} \\ 
\textsuperscript{$^*$ Equal contributions. A coin flip decides the order.} \\
\vspace{-1mm} 
\textsuperscript{$\dag$ Work done during an internship at Carnegie Mellon University.} 
} 

\abstract{
 Long-context large language models (LLMs) have recently shown strong performance in information retrieval and long-document QA. However, to tackle the most challenging intellectual problems, LLMs must reason effectively in long and complex contexts (e.g., frontier mathematical research). Studying how LLMs handle increasing reasoning complexity and context length is essential, yet existing benchmarks lack a solid basis for quantitative evaluation. Inspired by the abstraction of GSM-8K problems as computational graphs—and the ability to introduce noise by adding unnecessary nodes and edges—we develop a grade-school math problem generator capable of producing arithmetic problems with infinite difficulty and context length under fine-grained control. Using our newly synthesized \sysb benchmark, we comprehensively evaluate existing LLMs. We find a consistent sigmoid decline in reasoning performance as complexity increases, along with a systematic inference scaling trend: exponentially increasing inference computation yields only linear performance gains. These findings underscore the fundamental limitations of current long-context LLMs and the key challenges in scaling reasoning capabilities. Our \sysb benchmark provides a scalable and controllable testbed for systematically studying and advancing LLM reasoning in long and complex contexts.
}

\metadata[Github]{\url{https://github.com/Infini-AI-Lab/gsm_infinite/}}
\metadata[Website]{\url{https://infini-ai-lab.github.io/gsm_infinite/}} 

\begin{document}

\maketitle
\section{Introduction} 
\label{submission} 

\begin{wrapfigure}{r}{0.5\textwidth} 
    \includegraphics[width=\linewidth]{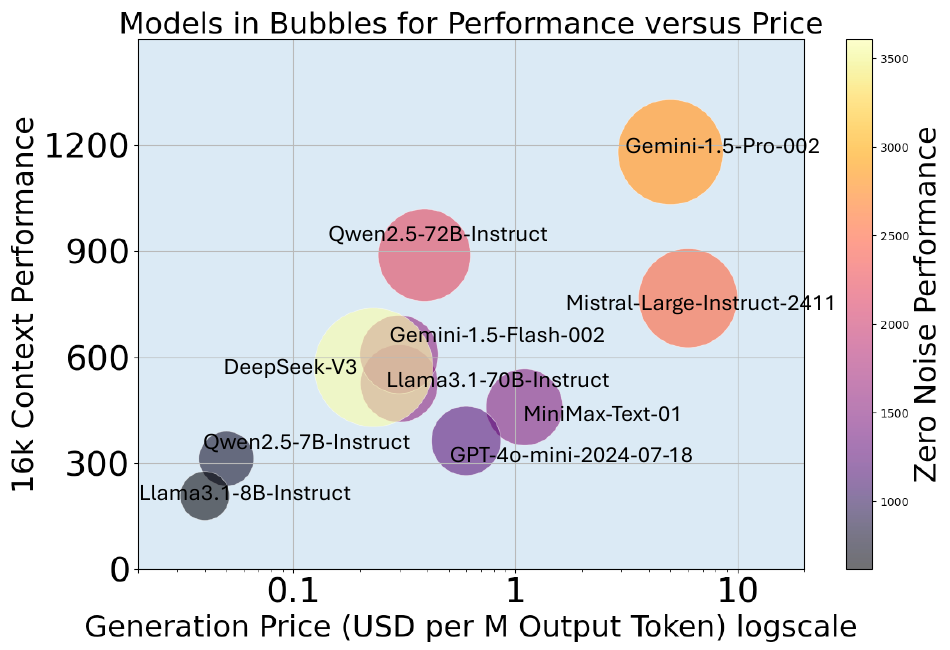} 
     \vspace{-5mm}
    \caption{Evaluation of 10 powerful LLMs on \sysb, comparing API generation cost (horizontal axis) with zero-context reasoning ability (vertical axis). Bubble size represents reasoning performance at a 16K context length.} 
    \vspace{-5mm} 
\end{wrapfigure} 

\begin{figure*}[t] 
    \centering
    \includegraphics[width=\textwidth]{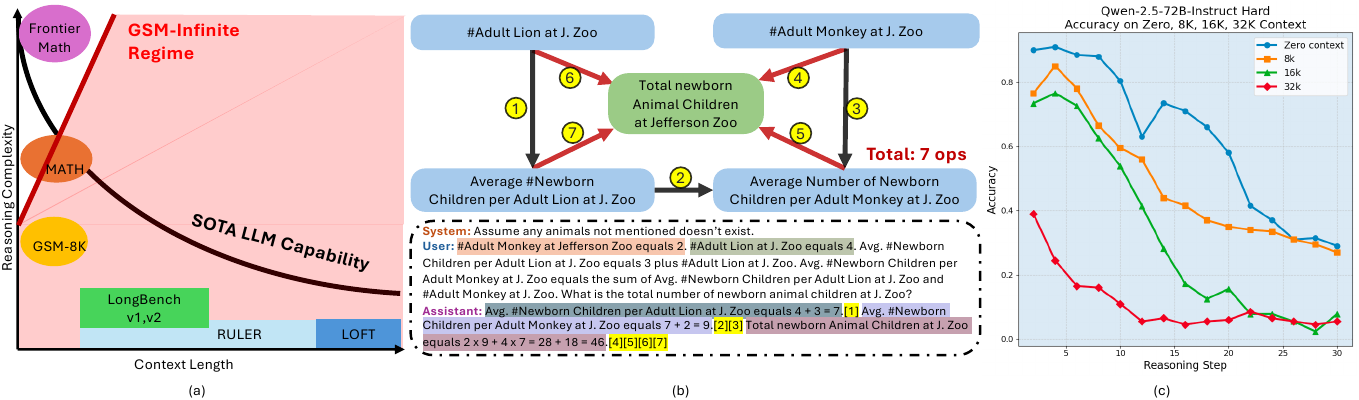} 
    \vspace{-5mm}
    \caption{(a) We position existing benchmarks across the Reasoning complexity versus context length plot. Reasoning datasets are usually of very short context. Existing long context benchmarks are usually low in reasoning complexity. Our task can cover any context length that the user so chooses and can generate infinite reasoning complexity. However, for high reasoning complexity, our task needs to use a longer context for problems. Our task is shown in Red. (b) A simplified example of our dataset-building process. We first generate an interconnected computational graph, and we then based on the graph, attach real-world context to it to formulate the problem statements. (c) Shows Qwen2.5-72B-Instruct Score decay across zero-context, 8K, 16K, and 32K.} 
    \label{figureone} 
    \vspace{-5mm}
\end{figure*} 

Recently, state-of-the-art long-context LLMs \citep{team2024gemini, minimax2025minimax01scalingfoundationmodels} have achieved astonishing performance in a tremendously long context, where \citet{team2024gemini} achieves near-perfect performance in 10M multimodal retrieval and long document QA. 
However, for long-context LLMs to contribute to cutting-edge mathematical and scientific discoveries or function as autonomous agents, they must be capable of processing dense, complex information and reason through multi-step tasks. 
For instance, Sir Andrew Wiles' proof \citep{wiles1995modular} of Fermat's Last Theorem in 1995 spans more than 88K highly compact tokens with deep logical connections, making context-level RAG~\citep{lewis2021retrievalaugmentedgenerationknowledgeintensivenlp} \textbf{insufficient} and highlighting the need for long-context LLMs. 
Therefore, it is crucial to benchmark and facilitate long-context LLMs for complex reasoning and high-density information processing.

\begin{figure*}
    \includegraphics[width=\textwidth]{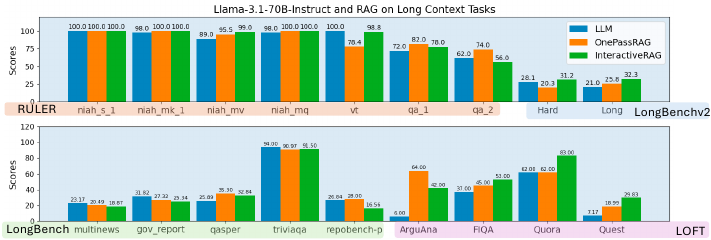} 
    \caption{Study of Llama3.1-70B-Instruct with Passive RAG (referred to as OnePassRAG) and Active RAG (referred to as InteractiveRAG) on popular long-context benchmarks: RULER (at 64K context length), LongBench ($>$8K), LongBenchV2, and LOFT (128K context length). RAG is under the 2048 retrieved token budget, and the decoder used for the RAG is Llama-3.1-70B-Instruct. RAGs generally have robust performance, on par with the corresponding LLMs, showing that previous long-context benchmarks are either too simple in reasoning complexity or contain detectable noise.} 
    \label{passiveplusinteractive} 
    \label{fig:3}
\end{figure*}

Although widely used, current long-context benchmarks do not fully capture the true potential of long-context LLMs~\citep{yu2024defense,li2024long,li2024retrieval}, making it challenging to measure their progress toward advanced intellectual agents. 
It is mainly due to the following three reasons: (1) \textbf{Low Complexity.} Many long-context benchmarks, such as LongBench \citep{bai2025longbenchv2deeperunderstanding, bai2024longbenchbilingualmultitaskbenchmark} and most tasks in RULER \citep{hsieh2024rulerwhatsrealcontext}, focus on retrieval or summarization, which involve low reasoning complexity. 
Similar to~\citep{yu2024defenserageralongcontext}, we found that simple context-level RAG achieves on-par or even better results than long-context LLMs (shown in~\cref{fig:3}). 
(2) \textbf{Detectable Noise.} Many tasks are innately short-context but are bloated into longer context through semantically irrelevant filler text (\citet{kuratov2024babilongtestinglimitsllms} and variable-tracing in \citet{hsieh2024rulerwhatsrealcontext}), which is easily distinguished by a retriever of context-level RAG. 
(3) \textbf{Low Resource.} While complex long-context tasks exist, such as long code completion \citep{loughridge2024dafnybenchbenchmarkformalsoftware}, they lack sufficient high-quality examples with adequate and verified annotation and labeling. This scarcity limits test diversity and fine-grained difficulty assessment, reducing their effectiveness in model evaluation. 

Ideally, a long-context reasoning benchmark should (1) offer controllable and \textbf{scalable complexity}, (2) incorporate \textbf{hard-to-distinguish noise}, and (3) support \textbf{infinite data} generation for continuous and adaptable evaluation. Inspired by \citet{delétang2023neuralnetworkschomskyhierarchy, ye2024physicslanguagemodels21}, we model reasoning problems as computational graphs attached with language semantics. By adjusting their structure and complexity, we gain fine-grained control over reasoning difficulty and enable infinite scaling. Instead of inserting semantically irrelevant filler text, noise is introduced as additional nodes and edges upon the core graph, strategically connected to existing nodes without affecting to the necessary reasoning steps for solving the tasks. This design enables the generation of arbitrarily long test examples, from which context-level RAG finds it hard to differentiate relevant information from noise. 

However, several technical challenges must be addressed to construct a practical benchmark. First, ensuring diverse reasoning patterns and operations - ideally providing a comprehensive coverage of GSM-8K - is crucial to enable fair and thorough evaluation. Second, computational graphs must be effectively translated into natural language that is both human- and LLM-readable, eliminating ambiguity from memorization and ensuring a clear focus on reasoning. 

We introduce \sysb, a long-context benchmarking framework that scales and controls reasoning complexity and noise through fine-grained manipulation of computational graphs, enabling their translation into diverse, human- and LLM-readable problems (Figure \ref{figureone}(b)). Specifically, 
\begin{itemize}[itemsep=0.0pt, topsep=0pt, leftmargin=*]
    \item in Section~\ref{sec:graphtoreasoning}, ~\ref{sec:noisecontruction}, we discuss how to control and scale reasoning complexity by manipulating computational graphs and how to insert noise. 
    \item Section~\ref{reverseproblem} discusses methods to ensure comprehensive coverage of reasoning patterns appeared in GSM-8K. 
    \item Section~\ref{attachlanguage} shows ways enabling the computational graphs to LLM-understandable natural language mapping. 
\end{itemize}

Figure \ref{figureone}(a) shows where our benchmark positions among existing benchmarks, demonstrating the effectiveness of \sysb for long-context reasoning evaluations.

We conduct a comprehensive evaluation of 17 state-of-the-art LLMs on zero-noise problems and 10 LLMs on various noise-injected tasks using \sysb. 
We inclusively covered a wide range of models, including both popular closed-source and open-source options, conventional transformers and hybrid architectures, models of varying sizes, and both reasoning and non-reasoning LLMs. 
In zero-noise settings, recent reasoning-optimized LLMs demonstrate substantial improvements over their non-reasoning counterparts. Notably, Deepseek-R1 \citep{deepseekai2025deepseekr1incentivizingreasoningcapability} achieves an average AUC score nearly four times higher than previous SOTA models. However, in noise-injected scenarios, LLMs exhibit varying degrees of performance degradation. Our analysis reveals several key observations:

\begin{itemize}[itemsep=0.0pt, topsep=0pt, leftmargin=*]
    \item \textbf{Decay with reasoning complexity:} LLM performance follows a strikingly consistent sigmoid or exponential decay as problem difficulty increases, highlighting fundamental limitations in scaling reasoning capabilities.  
     \item \textbf{Decay with noise:} Performance degradation intensifies as context length increases within the same difficulty level, while longer context brings sharper degradation of the LLM performance. 
    \item \textbf{LLM thinking patterns:} Models consistently perform better on forward-thinking tasks than on backward-thinking ones, suggesting a fundamental asymmetry in their reasoning strategies.  
    \item \textbf{Influence of Repeated sampling:} Repeated sampling~\citep{brown2024large} on \sysb reveals a clear pattern: performance improves linearly with increased inference steps but at an exponentially growing computation cost, underscoring inefficiencies in current LLM inference strategies.  
\end{itemize} 
 
\section{Related Work and Problem Statement} 
Despite the wide popularity of some existing long-context benchmarks, this section reveals and elaborates on the three key limitations: low complexity, detectable noise, and low resource or limited quantity of test examples. These three limitations make it extremely challenging to measure long-context LLMs' progress toward advanced intellectual agents using existing benchmarks. 

\textbf{Low Complexity.} A significant portion of long-context evaluation datasets, including RULER \citep{hsieh2024rulerwhatsrealcontext}, LongBench \citep{bai2024longbenchbilingualmultitaskbenchmark}, LongBench v2 \citep{bai2025longbenchv2deeperunderstanding}, and LOFT \citep{lee2024longcontextlanguagemodelssubsume}, primarily assess retrieval and summarization rather than complex reasoning. As shown in Figure \ref{passiveplusinteractive}, our experiments demonstrate that RAG systems achieve competitive results with Llama3.1-70B-Instruct across these datasets. Notably, RAG outperforms LLMs in retrieval-focused tasks (e.g., RULER, LOFT) and performs comparably in text summarization and QA (most LongBench tasks), as well as structured reasoning problems such as variable tracking (RULER-vt) and code completion (LongBench-repobench-p). RAG methods provide a strong baseline while being substantially more efficient.

\begin{wrapfigure}{r}{0.5\textwidth} 
    \includegraphics[width=0.5\textwidth]{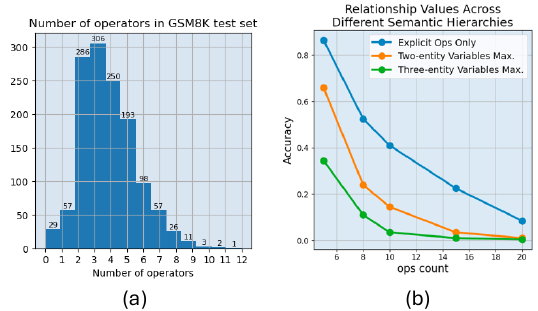} 
    \caption{(a) presents a conservative estimate for each problem difficulty in GSM-8K 1.3K test set. We evaluate the difficulty of the problems by the number of operations needed to get to the final answer. The op count ranges from 2 to 12, while most are around 3-4. (b) shows the Llama3.1-8B-Instruct performance across different semantics hierarchies, revealing the hidden reasoning difficulty innate in natural language.} 
    \label{gsm8kvariability} 
\end{wrapfigure} 

\begin{figure*}
    \includegraphics[width=\textwidth]{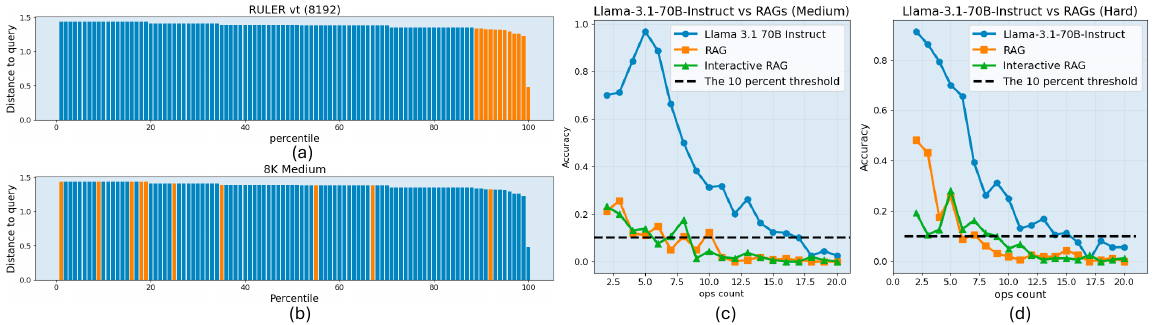} 
    \caption{RAG performance on our proposed long-context benchmarks. (a) studies retriever's behavior on the first 100 chunks of a random problem in vt from RULER with 8192 context length. The chunks that need to be retrieved to solve the problem are labeled in coral, while the noise is in blue. The chunks have retriever scores ranked from large (semantically far) to small (semantically close). Retriever locates the essential chunks with high precision, classifying all necessary chunks with the right side of the spectrum; (b) contrasts vt with our long-context benchmarks, showing that the retriever cannot locate precisely which chunk to retrieve. (c) and (d) display the performance of two RAG systems on our benchmark medium and hard tasks. (Figure best viewed in color)} 
    \label{mediumandhard} 
\end{figure*}

\textbf{Detectable Noise.} Many long-context benchmarks artificially extend short-context tasks by injecting extraneous text that does not contribute to solving the problem, allowing retrieval-based models to filter out noise effectively. In RULER's variable-tracing task with an 8192-token context, Llama3.1-70B-Instruct achieves 100\%, while OnePassRAG and InteractiveRAG reach 82.4\% and 98.4\%, respectively, despite using only a 2048-token retrieval budget. A detailed breakdown in Figure \ref{mediumandhard} (a) reveals that retrievers consistently identify and prioritize relevant information while disregarding injected noise. These findings indicate that existing long-context benchmarks do not adequately justify the need for expensive long-context LLMs, as RAG systems can effectively mitigate the impact of noise and achieve similar performance. 

\textbf{Low Resource.} Many high-quality reasoning tasks (Math and coding) heavily rely on human annotation and labeling and have test examples in limited quantity. Here we use GSM-8K \citep{cobbe2021trainingverifierssolvemath} as an example, shown in \cref{gsm8kvariability}(a). It is infeasible to extract subsets of examples with exact op at 8 for precise LLM evaluation due to the limited number of available cases—only 26 in total, with even fewer satisfying op $\geq$ 8. This scarcity makes meaningful evaluation impractical. Also, DafnyBench \citep{loughridge2024dafnybenchbenchmarkformalsoftware}, a high-quality long-context coding benchmark only contains 782 verified and deduped examples. 

\textbf{Problem Statement -} \textit{How can we develop a benchmark that contains sufficient problems at every fine-grained level of reasoning difficulty, from easy retrieval tasks to infinitely hard challenges, while providing infinitely customizable context length with high information density?} 
\label{lcllm} 

\section{Computational Graphs} 

\begin{figure*}
    \includegraphics[width=\textwidth]{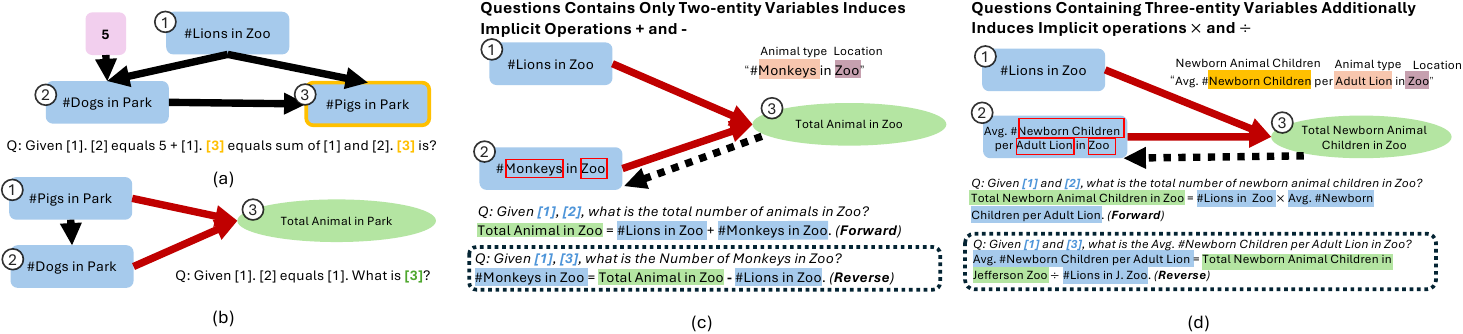} 
    \caption{Computational Graphs Illustration. (a) shows a simple computational graph, where every node and edge can be straightforwardly converted to a statement, the other way conversion works as well. (b) breaks down the essence of implicit operations and the abstract parameter constructions; here, an assumption provided in the system prompt is to assume the animals not mentioned in question don't exist; essentially, problems in natural language omit the two red arrows in the computational graph. (c) provides an example problem when all variables that appeared are ``two-entity" variables, where only implicit addition/subtraction can be generated. (d) contrasts (c) and shows an example that with additional ``three-entity" variables, the computation graph can also generate implicit multiplication/division. Both (c) and (d) also illustrate the design of reverse mode that specifically aims to generate implicit subtraction and division. (Figure best viewed in color)} 
    \label{mediumhard} 
\end{figure*} 

In this section, we detailed the construction of computational graphs, the key construct that enables \sysb generation of infinite quantities of arbitrary context length and reasoning complexity. Specifically, we explain in detail the potential mapping between reasoning problems and computational graphs in \Cref{sec:graphtoreasoning}, and in \Cref{sec:noisecontruction}, we propose to generate indistinguishable noise by strategically extending the computational graph. 

\subsection{Graph Construction to Build Reasoning Problems} 
\label{sec:graphtoreasoning}
After carefully studying GSM-8K problems, we draw the following crucial observations, which allow us to map a randomly generated computational graph to grade-school-level math reasoning problems that cover all possible operations and relationship types. 

\textbf{Mapping Explicit Ops to Computational Graphs - } From Every operation used in the GSM-8K is one of the four ``+”, ``$-$”, ``×”, and ``÷”. Consider the following example when operations are presented explicitly, ``Eggs cost twice as much as tomatoes, while tomatoes cost 1 dollar each.” These statements mention operations (``plus”, ``more”, ``times", etc.) can easily be abstracted out as a computational graph with variables, 
``dollar per egg” and ``dollar per tomato”, as nodes. There are two edges one pointing from "dollar per tomato" to "dollar per egg", while another one from a constant 2 to "dollar per tomato". Another similar example is shown in \cref{mediumhard} (a). 
Therefore, randomly generating a computational graph with different topology of edge connections will
lead to a new reasoning problem once the natural language
context are attached to the nodes of the graph. 

\textbf{Generating Implicit + using Computational Graphs - } On the other hand, the operations can also be presented implicitly hidden in natural language hierarchies. ``Mary earns 20 dollars in the morning, while she earns 25 dollars in the afternoon. How much total she earned that day?” Although the problem doesn’t explicitly mention addition, the solution has to sum up 20 and 25 to get 45. The reason is that natural language assumes a working day consists of morning and afternoon. Similarly, all four operations can be hidden in natural language hierarchies. Inspired by \citet{ye2024physicslanguagemodels21},
we adopt its construct of ``Abstract Parameters” and ``Instance Parameters” to construct computational graphs that facilitate the generation of problem statements containing the hidden operations. Essentially, the newly added constructs can be thought of as adding the ``total money” as a new node to the computational graph, which has two edges coming in, one from node ”Morning money” and the other one from node ``Afternoon money”. But when generating the problem, we omit the description of two edges pointing to the node ``total money on Friday”. We also illustrate it in \cref{mediumhard}(b). Following \cite{ye2024physicslanguagemodels21}, we connect the specific instance parameters to the abstract parameters using red edges. To reiterate, these edges are omitted, forcing the LLMs to use commonsense and inducing implicit operations. 

\textbf{Generating Implicit $\times$ using Computational Graphs -} every variables in the above-mentioned problem with the implicit ``+” operation are ”two-entity variables”. ``Morning Money” contains two entities, ``Morning” and ``Money”, where in the context,
``Money” is an attribute of ``Morning”. Same with ``After-
noon Money”. In fact, out of all the examples we manually examined in GSM-8K, the minimum number of entities in the variable name is two. However, if every variable in the problem are two-entity" variables, only generate implicit operations of + and - will be induced, but not $\times$ and $\div$. For a problem to contain implicit operations $\times$, problems must contain variables with more than two entities in its name. For example, ``Mary works 8 hours on Friday. Her hourly rate on
Friday is 10 dollars. How much she will earn on Friday
in total?” The variable ”money per hour” contains ``money”, ``hour”, and ``Friday” three entities, where
``money” is an attribute of ``hour”, while ``hour” is also an attribute of ``Friday”. We further contrast the above scenario in \cref{mediumhard}(c) and (d), we see that if every variable is two-entity or ``Animal" in ``Location", generating $\times$ isn't straightforward, but once we add in the third entity, or an attribute of animal, generation of $\times$ becomes natural. 

\textbf{Scaling up Reasoning Complexity and maintaining control with Computational Graph -} Our computational graph generator employs the abstract parameter construct to generate implicit operations and three-entity variables to represent multiplication operations. 
During the generation process of a problem, specifically, a query node is first sampled from the graph. The corresponding topological sort list—ending with the query—ensures the shortest solution path, serving as a measure of the problem's reasoning complexity. 
This approach enables the generation of a vast number of synthetic graphs. Furthermore, adjusting the number of variables provides a coarse control over complexity, which is refined through precise filtering to produce well-defined subsets of graphs that meet specific operation constraints.

\subsection{Noise Construction Using Computational Graphs} 
\label{sec:noisecontruction} 

\begin{wrapfigure}{r}{0.35\textwidth} 
    \includegraphics[width=0.35\textwidth]{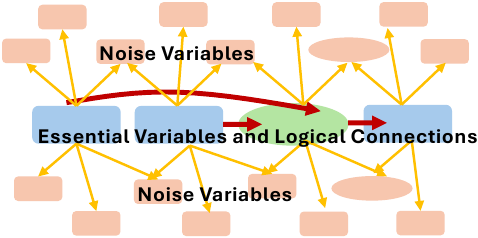} 
    \caption{The Illustration of Noise as Extension of Core Computational Graph.} 
    \label{noisytwo} 
\end{wrapfigure} 

\textbf{Spider Topology -} We observe that we can view noise as extending the computational graph to incorporate fake and unnecessary parameters and operators. However, two critical questions emerge. First, how to extend the computational graph without contaminating the original graph's solution and contaminations? We found that edges have to point outwards from the nodes in the original graph to the newly added noise nodes, essentially preventing the noise nodes from contributing to the core graph. Second, how to maximize the chance that RAG cannot retrieve the essential graph? It turns out interconnecting edges between newly added noise nodes won't contribute help detering RAG's retriever. We find out a simple trick works well: ensuring the majority of the added edges connect core nodes and the noise nodes contribute to a semantically close noise. We call this design Spider Topology as shown in Figure \ref{mediumhard}(c). 

We evaluated the resulting noise using two RAG systems presented in \ref{lcllm}. The results are shown in Figures \ref{mediumandhard}(c) and (d). Llama3.1-70B-Instruct achieves drastically stronger performance than the two RAG systems on 8K 2-entity problems and 3-entity problems. We also carried out the same study before on our data set setting 2, in Figure \ref{mediumandhard}(b). \textbf{We found that the RAG retriever now completely cannot distinguish which essential chunks from noise chunks, showing a clear contrast with vt tasks of the same context length in (a).}  
 
\section{\sysb} 
\label{gsminfinite} 

In this section, we present key techniques that enable the synthetic dataset to be diverse in operations in \cref{reverseproblem}, LLM-understandable, and enable the evaluation to be free from non-reasoning factors in \cref{attachlanguage}. Then, we present synthetic problem generators capable of generating grade-school math questions with arbitrary reasoning difficulty and context length. Thus, we generate a suite of benchmarks called \sysb detailed in \cref{fullbenchmark}. 

\subsection{Challenge 1: How to Generate Implicit $-$ and $\div$ Operations?} 
\label{reverseproblem} 
\textbf{Firstly, we review why the abstract-instance construct can only generate ``+" and $\times$ but not ``$-$" and $\div$.} 
The key limitation of \citet{ye2024physicslanguagemodels21} abstract parameters and instance parameter design is that it is only able to generate problems with solutions with the ``forward" and constructive ordering. Shown in Figure \ref{mediumhard} (a) and (b), the design dictates that the specific and detailed variables should be defined before a more abstract variable. For example, ``the number of Lions in Zoo" and ``the number of Monkeys in Zoo" have to be defined before ``Total Animal in Zoo" is defined. 
The ``forward" ordering leads to the inability to generate implicit '-' operations for 2-entity problems and implicit ``$\div$" operations for 3-entity problems that require the more abstract variables, e.g. ``Total Animal in Zoo", to be defined before a more specific variable, e.g. ``the number of Monkeys in Zoo". 

\textbf{To generate all four kinds of implicit operations, we introduce a ``reverse mode" to generate the computation graph.} 
Essentially, the graph construction still continues as before: starting with specific detailed variables and growing to incorporate more abstract variables. When it completes and we know all the values of nodes in the graph, we then randomly mask out specific initial low-level variables and force the solution to traverse in the reverse direction as in the "forward" ordering. 
We present the illustration of data generation in Figures \ref{mediumhard} (a) and (b) for the 2-entity and 3-entity, respectively. 
However, for 3-entity problems, it can result in quadratic equations leading to multiple possible solutions. We develop some techniques that effectively reduce the probability of the situation. Details are presented in \cref{sometips} of the Appendix. 
\label{reverse_problem} 

\subsection{Challenge 2: How to Ensure LLM-understandable Problem Generation?} 
\label{attachlanguage} 
\begin{wrapfigure}{r}{0.35\textwidth} 
    \includegraphics[width=0.35\textwidth]{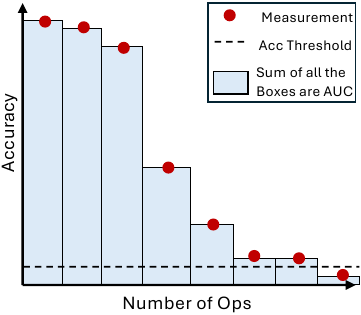} 
    \caption{Area Under Curve Metrics is Used to Compare between LLM Performance.} 
    \label{areacurve} 
\end{wrapfigure} 
Mapping computation graphs to natural language is critical for evaluating LLMs' reasoning capabilities. \textbf{To automate this process, we develop inter-swappable templates that enhance linguistic diversity while maintaining clarity.} Several key considerations inform our design. 

First, Certain syntactic forms, such as possessive constructions (e.g., A’s B), are straightforward to encode but can mislead LLMs due to their deviation from natural language. For example, South Zoo’s Penguin is restructured as Penguin in South Zoo, and South Zoo’s Adult Penguin’s Average Number of Newborn Children becomes Average Number of Newborn Children per Adult Penguin in South Zoo. Through extended trial and error, variable names with different entity numbers are \textbf{presented naturally} in all prepared templates. 

Second, the number of constraints applied to the random graph generation process is as little as possible, forcing templates to \textbf{enforce unit consistency} across two-entity and three-entity variables to enable assignment between these two. For instance, ``The average number of animal children per penguin in South Zoo" must share a unit with ``The number of penguins in South Zoo" to allow variable assignments.

Third, to ensure real-world knowledge doesn't confuse the LLM's decision, \textbf{we avoid specific real-world locations, people's names, and festival names} from appearing in the template. 
Based on these constricts, we propose three different templates that meet real-world templates: children-animal-zoo, teachers-school-district, and awards-movies-festival. We present in Appendix \ref{templates} an ablation study showing that three templates are consistent in overall performance with only minor fluctuations when evaluated using Llama-3.1-8B-Instruct. At each op, equal problems are tested for both constructive ordering (forward) and reverse ordering (reverse). 

\subsection{Benchmark Details} 
\label{fullbenchmark} 
\begin{table*}[ht] 
\centering 
\footnotesize 
\caption{18 selected models are evaluated on \sysb zero-noise benchmarks using Area-Under-Curve (AUC), which is computed by taking the Riemann Sum of accuracy versus op count from 2 to when the model accuracy drops below 5\%. We also present detailed statics of the first op number for the model to have an accuracy lower than 50\%, 10\%, and the average accuracy of the first 30 ops settings. Besides, we also highlight the \textcolor{yellow}{reasoning models}, \textcolor{green}{linear attention hybrid models}, and \textcolor{blue}{SSM hybrid models}. Due to space constraint, ``Mistral-Large-Instruct-2411" is shortened as ``Mistral-Large";``Claude-3.5-Sonnet" and ``Claude-3.5-Haiku" has version number 20241022; ``GPT-4o-2024-11-20" is shortened as ``GPT-4o" and ``GPT-4o-mini-2024-07-18" is shortened as ``GPT-4o-mini".} 
\begin{tabular}{@{}lccc|ccc|c@{}}
\toprule
\multirow{2}{*}{\textbf{Models}} & \multicolumn{3}{c|}{\textbf{Three Subtasks}} & \multicolumn{3}{c|}{\textbf{Detailed Statistics on Hard Subtask}} & \textbf{Score} \\
\cmidrule(lr){2-4} \cmidrule(lr){5-7}
 & \textbf{Symbolic} & \textbf{Medium} & \textbf{Hard} & \textbf{1st$<$50\% op} & \textbf{1st$<$10\% op} & \textbf{Avg. Acc op$\leq$30} & \textbf{Avg.$\uparrow$} \\
\midrule
\rowcolor{yellow!20} 
DeepSeek-R1 & 7280.0 & 9750.85 & 8573.8 & 100 & $>$130 & 0.9427 & 8534.88 \\ 
\rowcolor{yellow!20} 
GPT-o3-mini & 6690.0 & 8335.66 & 5769.96 & 70 & 110 & 0.9423 & 6931.88 \\
\rowcolor{yellow!20} 
GPT-o1-mini & 5060.0 & 6054.91 & 3738.43 & 50 & 90 & 0.8397 & 4951.11 \\
DeepSeek-V3 & 4310.0 & 4100.81 & 2407.86 & 24 & 55 & 0.6669 & 3606.22 \\
\rowcolor{yellow!20} 
QwQ-32B-preview & 3530.0 & 3205.75 & 1846.19 & 21 & 50 & 0.5403 & 2860.65 \\
Gemini-1.5-Pro-002 & 2547.0 & 3659.59 & 2318.28 & 26 & 45 & 0.6924 & 2841.62 \\
Claude-3.5-Sonnet & 2161.0 & 3281.8 & 2115.79 & 26 & 40 & 0.6758 & 2519.53 \\
Mistral-Large & 2332.5 & 2879.92 & 2310.49 & 24 & 50 & 0.6645 & 2507.64 \\
Qwen2.5-72B-Instruct & 2048.0 & 2496.81 & 2016.38 & 21 & 40 & 0.5433 & 2187.06 \\
GPT-4o & 2379.0 & 2457.37 & 1451.54 & 18 & 30 & 0.5064 & 2095.97 \\
Gemini-1.5-Flash-002 & 1970.0 & 1478.75 & 1274.25 & 13 & 30 & 0.4460 & 1574.33 \\
Llama3.1-70B-Instruct & 1769.0 & 1650.25 & 1205.25 & 15 & 30 & 0.4314 & 1541.50 \\
\rowcolor{palegreen!20} 
MiniMax-Text-01 & 1618.5 & 1712.64 & 1178.51 & 14 & 30 & 0.4213 & 1503.22 \\
GPT-4o-mini & 1389.0 & 1406.5 & 913.89 & 12 & 22 & 0.3094 & 1236.46 \\
Claude-3.5-Haiku & 897.0 & 1053.16 & 784.34 & 10 & 22 & 0.2910 & 911.50 \\
Qwen2.5-7B-Instruct & 786.95 & 886.75 & 618.5 & 7 & 19 & 0.2257 & 764.07 \\
Llama3.1-8B-Instruct & 462.0 & 786.5 & 606.5 & 6 & 17 & 0.2186 & 618.30 \\
\rowcolor{skyblue!20} 
Jamba-1.5-Large & 856.0 & 485.13 & 466.4 & 6 & 26 & 0.1828 & 602.51 \\
\bottomrule
\end{tabular} 
\label{zerocontextleaderb} 
\end{table*} 

With the synthetic problem generators detailed in Section \ref{gsminfinite}, we then use them to generate problems to build a suite of reasoning tasks with increasing complexity. For the brevity of reference, we refer to the generated problems with only explicit operations as ``Easy", the generated problems with 2-entity variables at maximum as ``Medium", and the generated problems with 3-entity variables at maximum as "Hard. 

Ideally, when evaluating an LLM, we want to evaluate all difficulty levels, from the most basic logic complexity to when it completely fails to solve any problem. For the Easy subset of problems, it usually leads to large operation counts for powerful LLMs. 
However, although complexity-wise not challenging, LLMs trained with internal COT tend to generate very long arguments, saturating their API output generation limit (4K for many models). Thus, we observe a sudden decay in accuracy in large ops, not because of LLMs' ability bottlenecks, but because of the above-mentioned nuance. Thus, we make a tweak to its problem: Instead of asking the LLM to find the value of one variable, we ask the LLM to find all the variables that have some value specified, effectively increasing the difficulty of the problem. 

\begin{wraptable}{r}{0.5\textwidth} 
\centering 
\caption{10 selected models are evaluated on \sysb Long Context benchmarks using Average AUC of Symbolic, Medium, and Hard. We evaluated models on 8K, 16K, and 32K context. Although our pipeline is capable of generating longer problems, the resource required to go further for larger models beyond our acceptance, while smaller models effectively has completely failed.} 
\footnotesize 
\begin{tabular}{@{}lcccc@{}}
\toprule
\textbf{Model} & \textbf{8K} & \textbf{16K} & \textbf{32K} & \textbf{Avg.$\uparrow$} \\
\midrule
Gemini-1.5-Pro-002 & 1182.43 & 896.31 & 812.96 & 963.9 \\
Qwen2.5-72B-Instruct & 927.33 & 681.53 & 563.65 & 724.17 \\
Mistral-Large & 914.49 & 563.73 & 319.21 & 599.14 \\
DeepSeek-V3 & 935.10 & 477.02 & 313.66 & 575.2 \\
Gemini-1.5-Flash-002 & 673.88  & 476.72 & 377.38 & 509.3 \\
Llama3.1-70B-Instruct & 479.00 & 394.50 & 355.5 & 409.67 \\
MiniMax-Text-01 & 481.32  & 359.56 & 325.95 & 388.94 \\
GPT-4o-mini & 401.00  & 337.81 & 275.63 & 338.15 \\
Qwen2.5-7B-Instruct & 248.00  & 211.50 & 196.17 & 218.56 \\
Llama3.1-8B-Instruct & 183.67  & 149.50 & 109.45 & 147.54 \\

\bottomrule
\end{tabular} 
\label{longcontextleaderb} 
\vspace{-5mm} 
\end{wraptable} 

For the modified Easy subset, we keep the generated problem in the most basic form: symbolic assignment. The typical problem statement then becomes "v1235 equals v1468 plus 1." Since the modified problem is not easier compared to Medium and Hard, we now call it "Symbolic". For Medium and Hard, we used all three templates and mixed the generated problems together to ensure diversity. For reporting LLM performance, we use \textbf{Area Under Curve (AUC)}, as shown in \cref{areacurve}, which is computing a Riemann sum over the LLM's performance in accuracy versus the number of operations from 2 to when its performance is lower than 5\%. 

We prepare zero-noise, 8K, 16K, and 32K in the benchmarks. The existing generation pipeline is capable of generating in $>$ 16M context, but the smaller 70B level models effectively failed in the 32K context already, while evaluating larger ones brings cost beyond our acceptance.

\section{Evaluation} 

In this section, we present comprehensive evaluations of various LLMs on \sysb. Specifically, the section is organized as follows: 
\begin{itemize}[leftmargin=*, itemsep=0pt, parsep=0pt, topsep=0pt] 
    \item \cref{leaderboardtwo} presents the complete comparison of LLMs on both zero-noise tasks and long-context tasks. Besides the main leaderboard, we further share four interesting findings. 
    \item \cref{sigmoid} reveals that the sigmoid function generally fits the LLM performance degradation to the increasing ops well. 
    \item \cref{forwardreverseeval} shows that LLMs generally and consistently perform forward problems better than reverse problems. (Defined in \ref{reverse_problem}) 
    \item \cref{longdegradation} discuss that various LLM performance decay over longer context and further ablation of the noise. 
    \item \cref{resam} shows that on \sysb, exponential increase in inference compute yields linear AUC gains. 
\end{itemize} 

\subsection{Leaderboard} 
\label{leaderboardtwo} 

\begin{wrapfigure}{r}{0.6\textwidth} 
    \centering
    \includegraphics[width=0.6\textwidth]{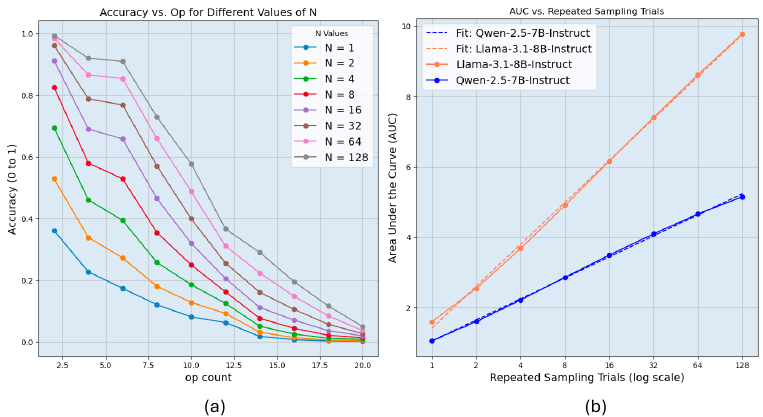} 
    \caption{(a) shows repeated sampling on zero-context Hard task with Qwen-2.5-7B-Instruct; (b) shows the AUC to repeated sampling number of trials. We show that for repeated sampling, exponentially increasing inference compute only leads to a linear increase in AUC improvement.} 
    \vspace{-7mm} 
    \label{aucrepeated} 
\end{wrapfigure} 

We evaluated 18 powerful LLMs on zero-noise problems, resulting in Table \ref{zerocontextleaderb}, while 10 models are evaluated on the long context as shown in Table \ref{longcontextleaderb}. 
From Table \ref{zerocontextleaderb}, we can see that the score separates these LLMs into clear groups. Reasoning models (R1 and o1-mini) are significantly ahead of the rest of non-reasoning LLMs. On the other hand, models with hybrid architecture aren't performing strong on the zero-noise pure reasoning benchmarks. MiniMax-Text-01 and Jamba both severely underperform compared to 70 B-level models. Also, similar things can be discussed when comparing 70B and 7B level models. In Table \ref{longcontextleaderb}, we see that the models show a very different decay pattern, while Gemini-1.5-Pro is significantly ahead of the rest of the models. 
Reasoning model evaluation remains too costly or slow for current long-context leaderboard participation. 

\begin{figure*}
    \includegraphics[width=\textwidth]{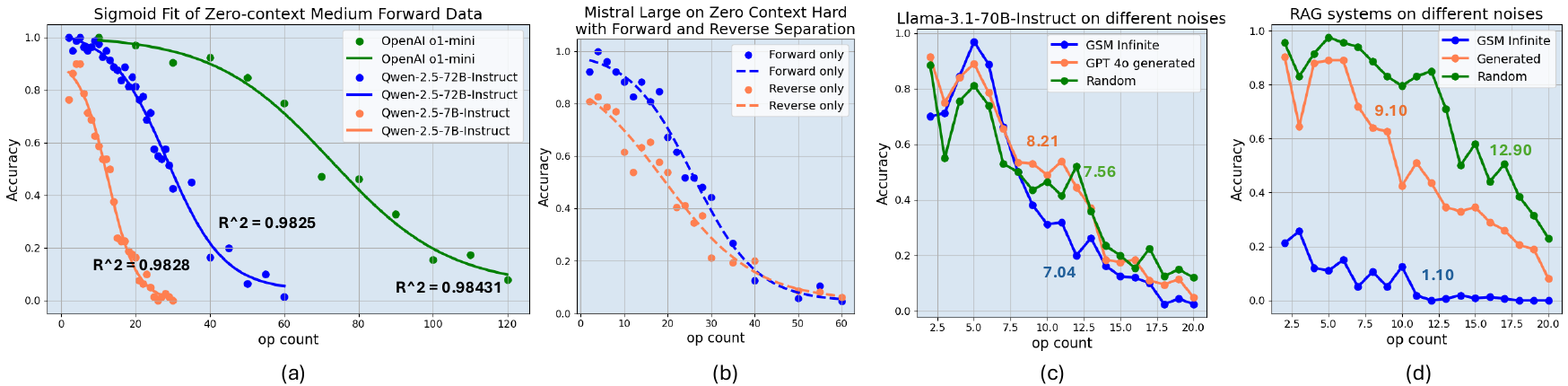} 
     \vspace{-6mm}
    \caption{(a) shows three different LLMs' behavior on \sysb benchmark zero-context Medium Forward, GPT o1-mini, Qwen2.5-72B-Instruct, and Qwen2.5-7B-Instruct. These models are drastically different in reasoning ability but have performance be modeled by sigmoid well, all with $R^2$ $>$0.98. (b) shows the gap between forward-thinking and reverse-thinking problems from Mistral Large on zero-context Hard. reverse-thinking problems are significantly harder and can be approximated by the sigmoid function that is essentially left-ward shifting from the forward sigmoid function. (c) and (d) presents RAG and corresponding LLMs' performance on different noises. Other than ours, RAG even improves performance.} 
    \label{xprmnt} 
    \vspace{-2mm}
\end{figure*} 

Later on, we examine the behavior of LLMs on \sysb and summarize four interesting findings uniquely enabled by our method of problem generation. 
\vspace{-3mm} 

\subsection{LLM Performance Degradation Can be Modeled Using Sigmoid Function} 
\label{sigmoid} 

Our construction of \sysb enables precise measurement of LLM performance across fine-grained difficulty levels. For each subtask, we observe a clear trend: LLM accuracy declines as the number of required operations increases. Surprisingly, most models exhibit a sigmoid-like performance decay, as shown in \ref{xprmnt} for forward problems (see Section \ref{reverse_problem} for definitions). Under the Medium subtask, LLM performance aligns remarkably well with a sigmoid function curve, with R² $>$ 0.98. The pattern is intuitive: at low operation counts, accuracy remains near 1.0; as complexity increases, performance first decays gradually, then drops sharply toward zero, where LLMs effectively fail to solve the problems. The score eventually stabilizes near zero. Interestingly, while LLMs vary in overall capability, they follow the same trend, differing primarily in the decay rate of the sigmoid function. 

\subsection{Reverse Problems are  Harder to Solve for LLMs} 
\label{forwardreverseeval} 

The generator of \sysb can generate both the "forward" and the "reverse" problems, which can be compared separately. Most LLMs perform worse in reverse problems than forward ones, shown in Figure \ref{xprmnt}(b) using Mistral-Large~\citep{jiang2023mistral}. For the Medium, only Jamba-1.5-Large out of a total of 18 LLMs evaluated do reverse problems better than forward problems, and the average difference in AUC is 604. 
For the Hard subtask, 5 out of 18 models have larger reverse problems AUC, Deepseek R1, Gemini-1.5-Pro-002, Qwen2.5-7B-Instruct~\cite{qwen2025qwen25technicalreport}, 4o-mini, and Jamba-1.5-Large~\citep{team2024jamba}. 
The average difference in AUC is 154. A detailed breakdown is listed in Appendix \ref{forwardreverse}. Besides, LLM performance on reverse problems can also be modeled by a sigmoid mapping.  Five more LLM plots are presented in Appendix \ref{forwardreverse}. 

\subsection{Long-context Degradation and Noise Ablation} 
\label{longdegradation} 
We evaluate LLM performance across increasing context lengths (0, 8K, 16K, 32K) and observe a consistent decline in performance as context length increases. Notably, models exhibit different decay patterns. 
We present results for 10 models across 3 subtasks, each with four curves representing different context lengths. All 30 plots are in Appendix \ref{picutrall}.

We conduct an ablation study on three noise types: \sysb (ours), LLM-generated, and random. For LLM-generated noise, we prompt GPT-4o to create a fake documentary-style commentary on random problems, occasionally introducing nonsensical variable mentions. For random noise, we follow Hsieh et al. (2024), using generic statements like "The sky is blue. The tree is green." We evaluate Llama3.1-70B-Instruct and a RAG system under all three noise types in an 8K context. Interestingly, RAG outperforms long-context LLMs on LLM-generated and random noise, effectively filtering irrelevant content. However, it fails to distinguish \sysb noise from essential problem statements.

\subsection{Limitations of Repeated Sampling} 
\label{resam} 
The construction of \sysb allows us to study techniques for Inference Scaling as well. Specifically, we study repeated sampling \citep{brown2024large,snell2024scalingllmtesttimecompute} and its effectiveness under different reasoning complexity. We study both Qwen2.5-7B-Instruct and Llama3.1-8B-Instruct with the best-of-N settings similar to \citet{brown2024largelanguagemonkeysscaling}. Interestingly, we find that repeated sampling seems to boost the performance the most for smaller op count subsets, and the benefit of repeated sampling diminishes gradually for larger op count subsets, Qwen2.5-7B-Instruct behavior is plotted in Figure \ref{aucrepeated}(a) with different repeated trial settings. 

Surprisingly, suppose we calculate the AUC score under every curve corresponding to each number of repeated trial settings. In that case, we find that the increment in the AUC score from two consecutive settings is close. If we plot the AUC score versus the number of repeated trial settings and take the log scale of the repeated trial N, the graph is linear, as shown in Figure \ref{aucrepeated}(b). The R-squared is greater than 0.99 for both Qwen2.5-72B-Instruct and Llama3.1-8B-Instruct, where Llama3.1-8B-Instruct. \textbf{Therefore, \sysb helps reveal that Repeated Sampling leads to linear AUC Improvement from exponentially increasing computation cost.} 
 
\section{Conclusion} 
Long-context LLMs have the potential to tackle complex, information-dense tasks requiring deep reasoning and coherent long-form generation. To advance their development and benchmarking, we introduce \sysb, a synthetic long-context reasoning benchmark generated entirely by a software-based system with fine-grained control over complexity and information density. Through extensive evaluations on \sysb, we uncover key insights to inform future LLM training and inference improvements.


\clearpage
\newpage
\bibliographystyle{assets/plainnat}
\bibliography{paper}

\clearpage
\newpage
\appendix 
\section*{Appendix Table of Contents}
We present the following table of contents to better traverse the Appendix. 

\begin{itemize}
    \item \hyperref[extrelatedworks]{Related Work}
    \begin{itemize}
        \item \hyperref[longcontextmodels]{Long-context Language Models}
        \item \hyperref[longcontextbenchmarks]{Long context benchmarks and tasks}
        \item \hyperref[reasoninglimitations]{Limitation of Existing Reasoning Tasks}
        \item \hyperref[synthesizeddatasets]{Synthesized Datasets for Long-context}
    \end{itemize} 
    \item \hyperref[limitation]{Benchmark Limitation} 
    \item \hyperref[setupr]{Detailed Experiment Setup}
    \item \hyperref[rag_full_result]{Full Result of RAG Experiments}
    \item \hyperref[illustrative_problems]{Illustrative Problems}
    \item \hyperref[forwardreverse]{Forward and Reverse Problems Breakdown}
    \item \hyperref[templates]{Ablation Study of Task Templates}
    \item \hyperref[picutrall]{Long-Context Degradation of Models}
\end{itemize} 

\section{Related Work} 
\label{extrelatedworks} 
\subsection{Long-context Language Models} 
\label{longcontextmodels} 
Various works related to the Long-context Language Model have been proposed. Flash attention\citep{dao2022flashattention}, Flash attention2\citep{dao2023flashattention2}, Ring attention\citep{liu2023ringattentionblockwisetransformers}, and Tree attention\citep{shyam2024treeattentiontopologyawaredecoding} significantly reduced the memory footprint and communication overhead for processing long context in engineering level across multiple nodes. Architectural level innovations such as sparse attentions represented by sliding window attention\citep{beltagy2020longformerlongdocumenttransformer}, are also widely used to reduce the overhead caused by the increasing sequence length. New training strategies, such as gradually extending the training context length in the final stages of pretraining have been applied to support a long context window\citep{dubey2024llama3herdmodels}.
\subsection{Long context benchmarks and tasks}
\label{longcontextbenchmarks} 
There have been quite a few works benchmarking long-context language models. Existing comprehensive benchmarks like $\infty$bench\citep{zhang2024inftybenchextendinglongcontext} cover realistic tasks including document QA, summary, and synthetic tasks including information retrieval, expression calculation, extending the context length in the benchmark to over 200k tokens. $\infty$bench\citep{zhang2024inftybenchextendinglongcontext} does have mathematical reasoning tasks, however the most relevant math.calc part seems to be too difficult for SOTA models to work out. Synthetic tasks often offer more control and are less affected by parametric knowledge in comparison with realistic tasks. One comprehensive synthetic benchmark is RULER\citep{hsieh2024ruler}, a synthetic benchmark with tasks including retrieval, variable tracking and so on, offering some controls over context length and task complexity. Experiments with various complexities were done, but it does not provide a quantitative analysis of complexity and context length on the correctness of the task, let alone isolate two separate patterns of performance decay. Other benchmarks usually focus on simple retrieval\citep{niah, liu2023lostmiddlelanguagemodels}, fact reasoning\citep{kuratov2024babilongtestinglimitsllms}, the impact of long context on natural language reasoning\citep{levy2024tasktokensimpactinput} and other real-world knowledge involved tasks. 
\subsection{Limitation of Existing Reasoning Tasks} 
\label{reasoninglimitations} 
Popular reasoning benchmarks are loose collections of human-made problems that naturally suffer from the following limitations. 
Firstly, the difficulty of problems within the same benchmark varies widely. 
We analyzed all 1.3K test problems in the GSM8K dataset, we plot the histogram in the number of operations in Figure \ref{gsm8kvariability}(a). The problem varied from 2 to over 12 following a skewed bell shape curve. 
This lack of fine-grained control makes it challenging to systematically evaluate models across incremental difficulty levels. 
Also, notice that the total number of problems is less than 10 for op $\geq$ 9, too little for stable evaluation. The lack of problem quantity on human-curated datasets eliminates the possibility of filtering out problems of each fine-grained difficulty level. 
Secondly, there is a significant difficulty gap between the benchmarks: GSM-8K focuses on middle school problems, MATH \citep{hendrycks2021measuringmathematicalproblemsolving} and AIME targets prospective university students, and Frontier Math challenges top-tier math graduate students. 
It is difficult to quantitatively determine the difference in problem difficulty between GSM-8K problems and MATH problems since MATH uses operations such as taking power or roots absent in GSM-8K. 
Similarly, it is not possible to determine the difference in complexity from MATH to Frontier Math. 
It is difficult to quantitatively model LLMs' performance degradation with the continuously increasing difficulty of the problem. 
Third, most of the existing problems have very short input prompts. On average, GSM-8K test set problems have a length of 59.96 tokens, while MATH test set problems have 67.37 tokens when using Llama 3.1 tokenizer. We have seen from \ref{lcllm} that the addition of irrelevant noise cannot meaningfully evaluate the ability to reason in a long context of LLMs. 
\subsection{Synthesized Datasets for long-context} 
\label{synthesizeddatasets} 
Synthesized tasks are simple to build and absolutely deterministic, data contamination safe, but highly effective in evaluating certain aspects of LLM performance. Its use in long-context benchmarks is profound. Needle-in-the-haystack \cite{kamradt2023needle}, a pioneering long-context synthesized task, now becomes the go-to task for evaluating LLM long-context retrieval ability. On the other hand, LLM reasoning benchmarks also see recent efforts in synthesized tasks. \cite{mirzadeh2024gsm} recently proposes to use build synthesized dataset upon GSM8K \cite{cobbe2021training} to study the robustness of LLM reasoning. \textbf{Part of our work draws a strong inspiration from a series of works (\cite{ye2024physicslanguagemodels21}, \cite{ye2024physicslanguagemodels22}) which systematically studies the intricacies of decoder transformers in solving grade-school level problems.} Following their footsteps, we carefully redesign the process of generating the problems so current LLMs can solve without training, and together with thoughtful steps in noise addition, we effectively construct reasoning benchmarks for the long-context community. 

\section{Benchmark Limitation} 
\label{limitation} 
Despite our careful construction and attention to LLM understandability, there are still several important limitations to the existing benchmark \sysb. First, our current benchmark only targets operations contained in the grades school math, so only +, $-$, $\times$, $\div$ are included. We are aware that Math \citep{hendrycks2021measuringmathematicalproblemsolving} includes more complex operators, which we are looking at incorporating into the next version of the benchmark. The key limitation it brings to our benchmark is shown in \cref{figureone}(a), to generate complex problems with high reasoning complexity, our generator needs to generate problems that are much longer than hard mathematic benchmarks, despite our limitless quantity and complexity scalability. Second, another key limitation of synthetic benchmarks is the lack of diversity in natural language. Natural Language reasoning tasks usually contain multiple ways to assign a variable value or to describe a relationship between multiple variables, which is extremely difficult to program in software without the LLMs in the loop. We deliberately aim to keep LLMs out of the loop for better scalability and counter the diversity by incorporating more than one template into the problem generation. Third, we implement multiple checks to ensure all of the essential steps in the core graph have positive values and have a number range of less than four figures, but for the current noise generation, we don't implement these checks, since they are redundant to the problems. However, we do notice that when processing through the noise, the LLMs sometimes are confused by the irregular noise variable values. In later version of the benchmark, we will address this issue and make sure the entire graph is consistent. 

\section{Detailed Experiment Setup} 
\subsection{RAG Experiment Setup} 
\label{setupr} 
The RAG system contains two components, the retriever and the decoder. For the retriever, we use all-mpnet-v2-base. For the decoder, we use Llama-3.1-70B-Instruct. The context retrieval budget for all problems is 2048. We employ two different RAG methods: passive and active RAGs. Passive RAG calls the retriever once before the generation of the decoder. The retriever computes the semantic similarity or distance between each chunk of context and the query sentence. These chunks in context are then ranked from closest (most semantic similar) to furthest (least semantic similar), and depending on the retrieved context, top-k chunks are retrieved. For our study, we used the L2 distance between context chunk embeddings and query embeddings. The decoder then takes the retrieved chunks as input and then outputs its response to the query. 

We use two types of RAG systems: Passive RAG that only calls the retriever once at the beginning to retrieve relevant context or Interactive RAG \cite{jiang2023activeretrievalaugmentedgeneration} in which the decoder decides when to retrieve, how many retrievals are needed, and generate a query for each retrieval. For the latter one, we restrict the decoder generation with only the latest retrieval content and its past generation. 

On the other hand, active RAG shows strong performance \cite{jiang2023activeretrievalaugmentedgeneration} especially for common sense reasoning tasks. In addition to the steps in passive RAGs, the decoder is allowed to initiate additional calls to the retriever to retrieve more context by generating new queries. We follow the state-of-the-art active RAG method FLARE \cite{jiang2023activeretrievalaugmentedgeneration}, which allows for 10 rounds of query, but restricts the LLM to only see its current round of retrieved context and its past rounds of generation to generate its full response.

\subsection{Repeated Sampling Experiment Setup}

\subsubsection{Procedure}
\begin{enumerate}
    \item \textbf{Oversampling Phase}
    \begin{itemize}
        \item Generate 256 samples per task with temperature $T=1.0$
        \item Use fixed random seeds for reproducibility
    \end{itemize}
    
    \item \textbf{Accuracy Calculation}
    \begin{itemize}
        \item Compute per-task empirical accuracy:
        \begin{equation}
            p_{\text{task}} = \frac{\text{\# Correct Samples}}{256}
        \end{equation}
        
        \item Estimate accuracy for N samples:
        \begin{equation}
            \text{Acc}_{\text{task}} = 1 - (1 - p_{\text{task}})^{N}
        \end{equation}
    \end{itemize}
    
    \item \textbf{Aggregation}
    \begin{itemize}
        \item Average results across 80 tasks:
        \begin{equation}
            \text{Final Accuracy} = \frac{1}{80} \sum_{i=1}^{80} \text{Acc}_{\text{task}_i}
        \end{equation}
    \end{itemize}
\end{enumerate}

\subsubsection{Rationale}
\begin{itemize}
    \item \textbf{Oversampling}: 256 samples reduces variance in estimating $p_{\text{task}}$ compared to using 128 samples directly
    \item \textbf{Probability Formula}: Models cumulative success probability:
    \begin{equation*}
        P(\geq\text{1 correct in }k\text{ trials}) = 1 - (1 - p)^k
    \end{equation*}
    \item \textbf{Task Count}: 80 tasks per op provide stable statistics while remaining computationally feasible
\end{itemize}

\subsection{Some Small Implementation Tips to Avoid Quadratic (or higher-order) Equations and Multiple Possible Solutions} 
\label{sometips} 
The quadratic equation gives rise to the following particular situation. For solving reverse mode questions, one easy way is to assign the variable asked as X, perform symbolic operations carrying with X, and go through the entire computation graph to close the circle and formulate an equation. Using that equation would easily solve for X. However, sometimes the query variable to too deep in the chain, and it might assign both variables Y and Z, which are later multiplied. Then, two linear expressions are multiplied to get quadratic operations, and as op becomes larger, it can be more severe than quadratic. 

To alleviate the situation, we found that the following technique is empirically helpful. When generating the problem statement, we first locate the variable that has the largest number of ``entities" in the variable name and make sure that that one is the end of the topology sort list during the forward mode generation. Then, when switching to reverse mode, we locate the last possible initialized variable, so as close to the end of the sort list as possible, and assign this variable as a query, effectively reducing the chance of its dependents multiplying each other. The method can reduce the chance of quadratic or higher-order equations happening. However, if we still encounter the cases even when the above procedure is in place, we discard the generated example and ask the generator to generate again. 

\newpage 

\section{Full Result of RAG experiments}
\label{rag_full_result}
Here, we present the full result of the RAG experiment for further analysis. We retrieve 2048 tokens for each RAG retrieval.
\subsection{RULER}
\begin{longtable}{l| c c c c c c c c |c}
\hline
\textbf{Models} & \textbf{s1} & \textbf{s2} & \textbf{s3} & \textbf{mk1} & \textbf{mk2} & \textbf{mk3} & \textbf{mv} & \textbf{mq} & \textbf{Context Length} \\ \hline
\endhead

Llama3.1-70B-Instruct & 100 & 100 & 100 & 100 & 100 & 100 & 100 & 100 & 8k \\ 
OnePass RAG & 100 & 100 & 100 & 100 & 100 & 96 & 98.5 & 100 & 8k \\ 
Interactive RAG & 100 & 100 & 100 & 100 & 100 & 98 & 99 & 99 & 8k \\ 
Llama3.1-70B-Instruct & 100 & 100 & 100 & 100 & 98 & 100 & 98 & 100 & 32k \\ 
OnePass RAG & 100 & 100 & 100 & 98 & 100 & 72 & 99.5 & 97.5 & 32k \\ 
Interactive RAG & 100 & 100 & 100 & 98 & 96 & 96 & 98.5 & 98.5 & 32k \\ 
Llama3.1-70B-Instruct & 100 & 100 & 100 & 98 & 96 & 100 & 89 & 98 & 64k \\ 
OnePass RAG & 100 & 100 & 100 & 100 & 100 & 56 & 95.5 & 100 & 64k \\
Interactive RAG & 100 & 100 & 100 & 100 & 96 & 98 & 99 & 100 & 64k \\ \hline

\caption{RAG vs Model (RULER NIAH)}
\label{tab:rag_performance_ruler_part1}
\end{longtable}

\begin{longtable}{l|c c c c c|c}
\hline
\textbf{Models} & \textbf{vt} & \textbf{cwe} & \textbf{fwe} & \textbf{qa1} & \textbf{qa2} & \textbf{Context Length} \\ \hline
\endhead

Llama3.1-70B-Instruct & 100 & 100 & 96.67 & 84 & 74 & 8k \\ 
OnePass RAG & 82.4 & 14.8 & 97.33 & 86 & 86 & 8k \\ 
Interactive RAG & 98.4 & 31.2 & 79.33 & 80 & 68 & 8k \\ 
Llama3.1-70B-Instruct & 100 & 95.2 & 97.33 & 80 & 66 & 32k \\ 
OnePass RAG & 86 & 5.2 & 92 & 84 & 74 & 32k \\ 
Interactive RAG & 98 & 7.6 & 80 & 78 & 64 & 32k \\ 
Llama3.1-70B-Instruct & 100 & 6.2 & 95.33 & 72 & 62 & 64k \\ 
OnePass RAG & 78.4 & 1.2 & 88.67 & 82 & 74 & 64k \\ 
Interactive RAG & 98.8 & 2.6 & 72 & 78 & 56 & 64k \\ \hline

\caption{RAG vs Model (RULER other subsets)}
\label{tab:rag_performance_ruler_part2}
\end{longtable}
\textbf{Comments}: s1-3 = niah\_single\_1-3, mk1-3 = niah\_multikey\_1-3, mv = niah\_multivalue, mq = niah\_multiquery

\subsection{LongBench V2}
\begin{longtable}{l|c|c c c c}
\hline
\textbf{Tasks} & \textbf{Overall} & \textbf{Easy} & \textbf{Hard} & \textbf{Short} & \textbf{Long} \\
\hline
\endhead
Llama3.1-70B-Instruct & 30 & 33.3 & 28.1 & 44.7 & 21 \\
OnePassRAG  & 25 & 33.3 & 20.3 & 23.7 & 25.8 \\
InteractiveRAG  & 33 & 36.1 & 31.2 & 34.2 & 32.3 \\
\hline
\caption{RAG vs Model (LongBench V2)}
\label{tab:rag_performance_longbenchv2}
\end{longtable}

\subsection{LongBench}

\begin{longtable}{l|c c c}
\hline
\textbf{Tasks} & \textbf{passage\_count} & \textbf{hotpot-qa} & \textbf{samsum} \\
\hline \endhead
Llama3.1-70B-Instruct & 36.0,36.0,32.0 & 58.87,71.22,76.44 & 28.88,35.95,41.48 \\
OnePassRAG  & 0.0,0.0,0.0 & 65.04,61.59,63.05 & 31.63,23.28,26.84 \\
InteractiveRAG  & 27.0,14.0,6.0 & 61.86,51.0,55.94 & 24.83,20.21,23.81 \\
\hline
\caption{RAG vs Model (LongBench) - Part 1}
\label{tab:longbench_part1}
\end{longtable}

\begin{longtable}{l|c c c}
\hline
\textbf{Tasks} & \textbf{multi-news} & \textbf{multifieldqa\_en} & \textbf{gov\_report} \\
\hline \endhead
Llama3.1-70B-Instruct & 27.71,24.81,23.17 & 57.31,51.83,64.98 & 34.94,34.97,31.82 \\
OnePassRAG  & 26.85,22.72,20.49 & 51.69,48.66,55.85 & 32.6,30.67,27.32 \\
InteractiveRAG  & 24.64,19.99,18.87 & 47.71,42.98,58.45 & 29.91,27.02,25.34 \\
\hline
\caption{RAG vs Model (LongBench) - Part 2}
\label{tab:longbench_part2}
\end{longtable}

\begin{longtable}{l|c c c}
\hline
\textbf{Tasks} & \textbf{qasper} & \textbf{passage\_retrieval\_en} & \textbf{2wikimqa} \\
\hline \endhead
Llama3.1-70B-Instruct & 50.3,46.5,25.89 & 100.0,100.0,100.0 & 74.93,64.37,59.6 \\
OnePassRAG  & 45.73,43.5,35.3 & 72.0,72.0,79.0 & 67.97,59.64,48.4 \\
InteractiveRAG  & 43.8,37.9,32.84 & 91.0,90.0,85.33 & 46.04,43.46,38.8 \\
\hline
\caption{RAG vs Model (LongBench) - Part 3}
\label{tab:longbench_part3}
\end{longtable}

\begin{longtable}{l|c c c c}
\hline
\textbf{Tasks} & \textbf{triviaqa} & \textbf{trec} & \textbf{lcc} & \textbf{repobench-p} \\
\hline \endhead
Llama3.1-70B-Instruct & 82.0,93.6,94.0 & 48.0,12.0,12.0 & 50.14,55.0,50.04 & 29.98,27.82,26.84 \\
OnePassRAG  & 92.13,89.46,90.97 & 47.0,56.0,53.0 & 19.92,14.5,18.36 & 34.76,33.62,28.0 \\
InteractiveRAG  & 88.11,92.32,91.5 & 56.0,57.0,52.0 & 24.26,23.26,22.57 & 14.97,17.15,16.56 \\
\hline
\caption{RAG vs Model (LongBench) - Part 4}
\label{tab:longbench_part4}
\end{longtable}
\textbf{Comments}: The 3 data separated by commas are subsets of 0-4k, 4-8k,8k+ respectively

\subsection{LOFT}
\begin{longtable}{l|c c c c c c c}
\hline
\textbf{Tasks} & \textbf{ArguAna} & \textbf{FEVER} & \textbf{FIQA} & \textbf{MS MARCO} & \textbf{NQ} & \textbf{Quora} & \textbf{SciFact} \\
\hline \endhead 
Llama3.1-70B-Instruct & 0.06 & 0.78 & 0.37 & 0.67 & 0.84 & 0.62 & 0.59 \\
OnePassRAG  & 0.64 & 0.88 & 0.45 & 0.77 & 0.86 & 0.62 & 0.64 \\
InteractiveRAG  & 0.42 & 0.73 & 0.53 & 0.69 & 0.76 & 0.83 & 0.87 \\
\hline
\caption{RAG vs Model (LOFT) - Part 1}
\end{longtable}

\begin{longtable}{l|c c c c c c}
\hline
\textbf{Tasks} & \textbf{Touché-2020} & \textbf{HotPotQA} & \textbf{MuSiQue} & \textbf{QAMPARI} & \textbf{QUEST}  \\
\hline \endhead
Llama3.1-70B-Instruct & 0.4411 & 0.37 & 0.2 & 0.024 & 0.07166 \\
OnePassRAG  & 0.2529 & 0.455 & 0.2383 & 0.1559 & 0.1899  \\
InteractiveRAG  & 0.79 & 0.29 & 0.13 & 0.1539 & 0.2983 \\
\hline
\caption{RAG vs Model (LOFT) - Part 2}
\label{tab:your_label}
\end{longtable}
\textbf{Comments}: Due to LOFT's limited prompt availability (no official releases for 32k/1M contexts), we conducted experiments solely on 128k context lengths. We observed that LOFT’s document ID retrieval tasks inherently require document identifiers that aren’t captured in standard RAG-retrieved chunks. To address this, we implemented a lightweight modification: appending document ID tags to each context chunk during retrieval. This allows the RAG system to infer the correct document ID when retrieving relevant content, resolving the task-specific limitation without altering core RAG functionality. 
\newpage

\section{Illustrative Problems} 
\label{illustrative_problems} 
This section presents one representative problem from each subset (Symbolic, Medium, and Hard) defined in the appendix. These examples illustrate the variations within the benchmark. see Table \ref{tab:illustrative_problems}

\begin{table}[h]
\caption{Illustrative Problems from Each Subset}
\label{tab:illustrative_problems} 
\begin{tabular}{llll}
\toprule
\multicolumn{4}{l}{Three example problems one for each subtask} \\ 
\midrule
Problem & \multicolumn{3}{p{0.8\textwidth}}{\vspace{-0.5\baselineskip} 
\begin{itemize}
\item \textbf{Symbolic (op=5):} $<$context$>$\textbackslash nassign V705804 = V437110 + 1. assign V986916 = V705804. assign V873548 = 6. assign V684196 = V873548. assign V437110 = V873548.\textbackslash n $<$/context$>$ \textbackslash n\textbackslash nThe context contains relationships between variables. These relationships are independent mathematical equations that are all satisfied simultaneously.\textbackslash n Using only these relationships, determine which variables (if any) from which values can be derived are equal to 7.\textbackslash nShow your step-by-step reasoning and calculations, and then conclude your final answer in a sentence. \textbf{Answer}: V705804,V986916.
\item \textbf{Medium (op=5):} Problem: The number of adult owl in Bundle Ranch equals 2 times the number of adult eagle in Bundle Ranch. The number of adult eagle in Hamilton Farm equals the difference between the total number of adult animals in Bundle Ranch and the number of adult eagle in Bundle Ranch. The number of adult owl in Hamilton Farm equals 4 times the number of adult owl in Bundle Ranch. The number of adult eagle in Bundle Ranch equals 3.  Question: What is the total number of adult animals in Bundle Ranch?
\textbf{Answer}: 9.
\item \textbf{Hard (op=5):} The average number of newborn children per adult blue jay in Bundle Ranch equals 2. The number of adult parrot in Bundle Ranch equals 2. The number of adult blue jay in Bundle Ranch equals 2 times the average number of newborn children per adult blue jay in Bundle Ranch. The number of adult eagle in Bundle Ranch equals 2 times the average number of newborn children per adult blue jay in Bundle Ranch. The number of adult parrot in South Zoo equals 4 times the sum of the average number of newborn children per adult eagle in Hamilton Farm, the number of adult eagle in Hamilton Farm, and the average number of newborn children per adult eagle in Hamilton Farm. The average number of newborn children per adult eagle in Hamilton Farm equals the number of adult eagle in Bundle Ranch. The number of adult eagle in Hamilton Farm equals 3. The average number of newborn children per adult parrot in Bundle Ranch equals the total number of adult animals in Hamilton Farm. The number of adult eagle in South Zoo equals 1. The average number of newborn children per adult parrot in South Zoo equals the average number of newborn children per adult parrot in Bundle Ranch. The average number of newborn children per adult eagle in Bundle Ranch equals 3 plus the average number of newborn children per adult parrot in Bundle Ranch. The average number of newborn children per adult eagle in South Zoo equals the sum of the number of adult blue jay in Bundle Ranch, the average number of newborn children per adult blue jay in Bundle Ranch, the average number of newborn children per adult parrot in Bundle Ranch, and the number of adult parrot in Bundle Ranch.  Question: What is the average number of newborn children per adult eagle in Bundle Ranch?
\textbf{Answer}: 6.
\end{itemize}
}\\ 
\bottomrule
\end{tabular}
\end{table}

\newpage
\section{Forward and Reverse Problems Breakdown} 
\label{forwardreverse} 

\begin{longtable}{|l|c|c|c|} 
\hline
\textbf{Models} & \textbf{Forward Problem} & \textbf{Reverse Problem} & \textbf{Forward AUC - Reverse AUC} \\ \hline
Llama3.1-70B-Instruct & 2100.625000 & 1283.750000 & 816.875000 \\ \hline
GPT-4o-mini & 1529.725400 & 1267.579000 & 262.146400 \\ \hline
\rowcolor{yellow!20} 
Jamba-1.5-Large & 390.380000 & 624.980000 & -234.600000 \\ \hline
GPT-4o & 3073.997375 & 1952.816875 & 1121.180500 \\ \hline
Mistral-Large & 3468.234100 & 2431.732450 & 1036.501650 \\ \hline
Llama3.1-8B-Instruct & 1030.000000 & 563.125000 & 466.875000 \\ \hline
Claude-3.5-Sonnet & 3653.830050 & 3158.657850 & 495.172200 \\ \hline
Qwen2.5-72B-Instruct & 2889.375000 & 2141.250000 & 748.125000 \\ \hline
Qwen2.5-7B-Instruct & 995.625000 & 833.125000 & 162.500000 \\ \hline
o1-mini & 6517.510550 & 5592.307100 & 925.203450 \\ \hline
Gemini-1.5-Flash-002 & 1889.375000 & 1153.750000 & 735.625000 \\ \hline
Claude-3.5-Haiku & 1234.620000 & 873.100000 & 361.520000 \\ \hline
Llama-3.1-405B-Instruct & 1781.400000 & 981.250000 & 800.150000 \\ \hline
DeepSeek-V3 & 4613.125000 & 3713.125000 & 900.000000 \\ \hline
Gemini-1.5-Pro-002 & 4204.564075 & 3160.574950 & 1043.989125 \\ \hline
DeepSeek-R1 & 9764.950000 & 9750.950000 & 14.000000 \\ \hline
MiniMax-Text-01 & 2148.071300 & 1539.415650 & 608.655650 \\ \hline
QwQ-32B-Preview & 3530.000000 & 2846.250000 & 683.750000 \\ \hline 
\caption{Medium Difference in AUC in Forward Problems and Reverse Problems} 
\end{longtable} 

\begin{longtable}{|l|c|c|c|}
\hline
\textbf{Models} & \textbf{Forward Problem} & \textbf{Reverse Problem} & \textbf{Forward AUC - Reverse AUC} \\ \hline
Claude-3.5-Haiku & 819.240000 & 776.900000 & 42.340000 \\ \hline
Llama3.1-70B-Instruct & 1314.375000 & 1098.750000 & 215.625000 \\ \hline
Gemini-1.5-Flash-002 & 1341.250000 & 1219.375000 & 121.875000 \\ \hline
MiniMax-Text-01 & 1360.555000 & 1034.625000 & 325.930000 \\ \hline
\rowcolor{yellow!20}
DeepSeek-R1 & 8444.500000 & 8756.950000 & -312.450000 \\ \hline
o1-mini & 3831.381000 & 3645.474200 & 185.906800 \\ \hline
\rowcolor{yellow!20}
Gemini-1.5-Pro-002 & 2255.732025 & 2444.270375 & -188.538350 \\ \hline
DeepSeek-V3 & 2725.085000 & 2109.560000 & 615.525000 \\ \hline
\rowcolor{yellow!20}
Qwen2.5-7B-Instruct & 625.625000 & 630.625000 & -5.000000 \\ \hline
GPT-4o & 1592.280000 & 1311.560000 & 280.720000 \\ \hline
Llama3.1-8B-Instruct & 759.375000 & 460.625000 & 298.750000 \\ \hline
Qwen2.5-72B-Instruct & 2196.875000 & 1895.000000 & 301.875000 \\ \hline
Claude-3.5-Sonnet & 2242.309950 & 1999.998100 & 242.311850 \\ \hline
\rowcolor{yellow!20}
GPT-4o-mini & 858.400000 & 873.310000 & -14.910000 \\ \hline
QwQ-32B-Preview & 1878.750000 & 1855.625000 & 23.125000 \\ \hline
Mistral-Large & 2570.940500 & 2018.469000 & 552.471500 \\ \hline
Llama-3.1-405B-Instruct & 1215.000000 & 743.750000 & 471.250000 \\ \hline
\rowcolor{yellow!20}
Jamba-1.5-Large & 274.980000 & 699.990000 & -425.010000 \\ \hline 
\caption{Hard Difference in AUC in Forward Problems and Reverse Problems} 
\end{longtable} 

\begin{figure}[h]
  \centering
  \begin{subfigure}[b]{0.3\textwidth}
    \includegraphics[width=\textwidth]{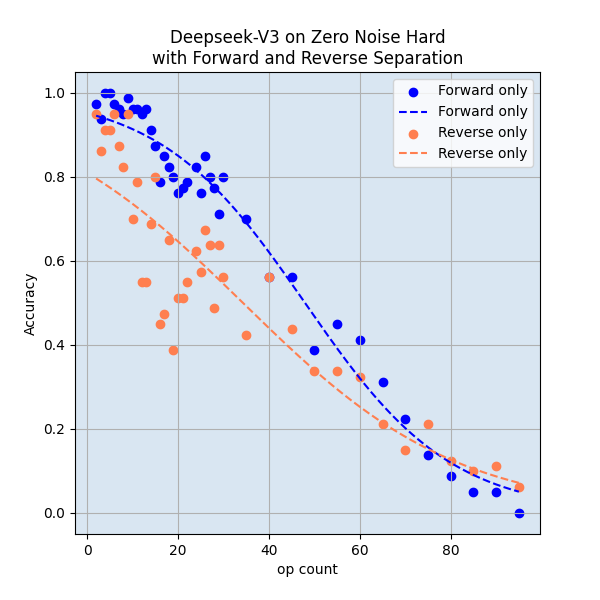}
    \caption{DeepSeek}
  \end{subfigure}
  \hfill
  \begin{subfigure}[b]{0.3\textwidth}
    \includegraphics[width=\textwidth]{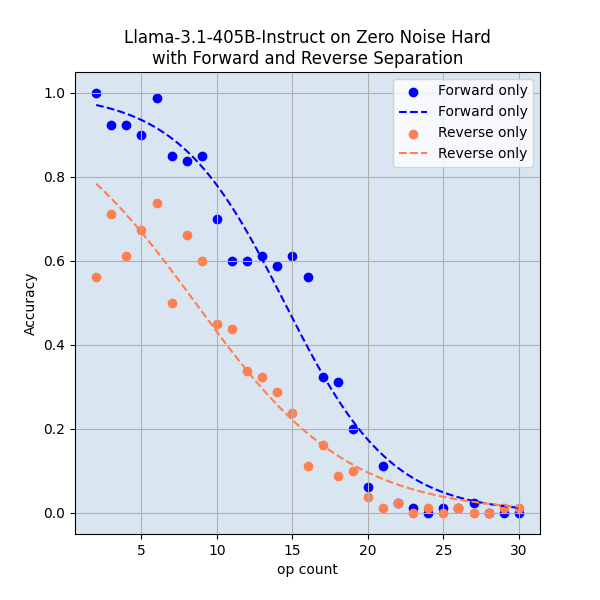}
    \caption{LLaMA-405B}
  \end{subfigure}
  \hfill
  \begin{subfigure}[b]{0.3\textwidth}
    \includegraphics[width=\textwidth]{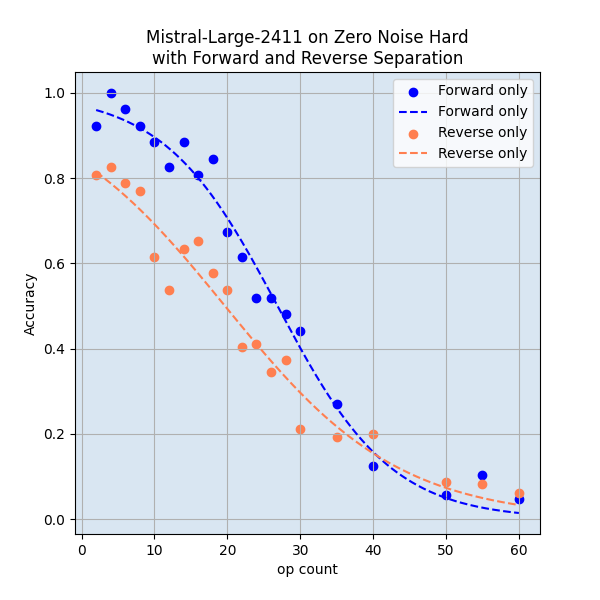}
    \caption{Mistral}
  \end{subfigure}
  \caption{Comparison of different models}
  \label{fig:models}
\end{figure} 

\begin{figure}[h] 
  \centering
  \begin{subfigure}[b]{0.3\textwidth}
    \includegraphics[width=\textwidth]{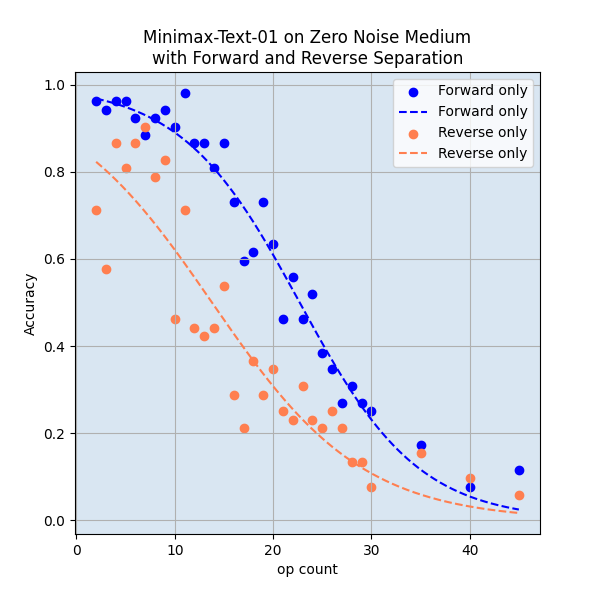} 
  \end{subfigure} 
  \hspace{0.05\textwidth} 
  \begin{subfigure}[b]{0.3\textwidth} 
    \includegraphics[width=\textwidth]{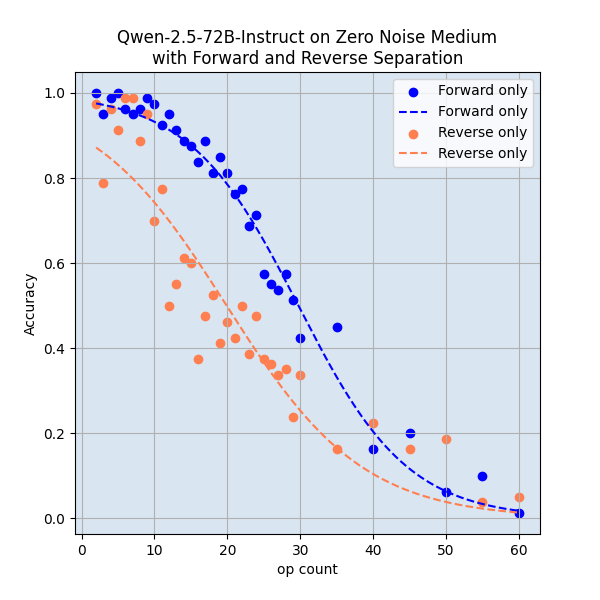} 
  \end{subfigure} 
  \caption{Comparison between forward and reverse}
  \label{fig:forwardreverse} 
\end{figure} 

\newpage

\section{Ablation Study of Task Templates}
\label{templates} 
\begin{figure}[h!]
  \includegraphics[width=1.0\textwidth]{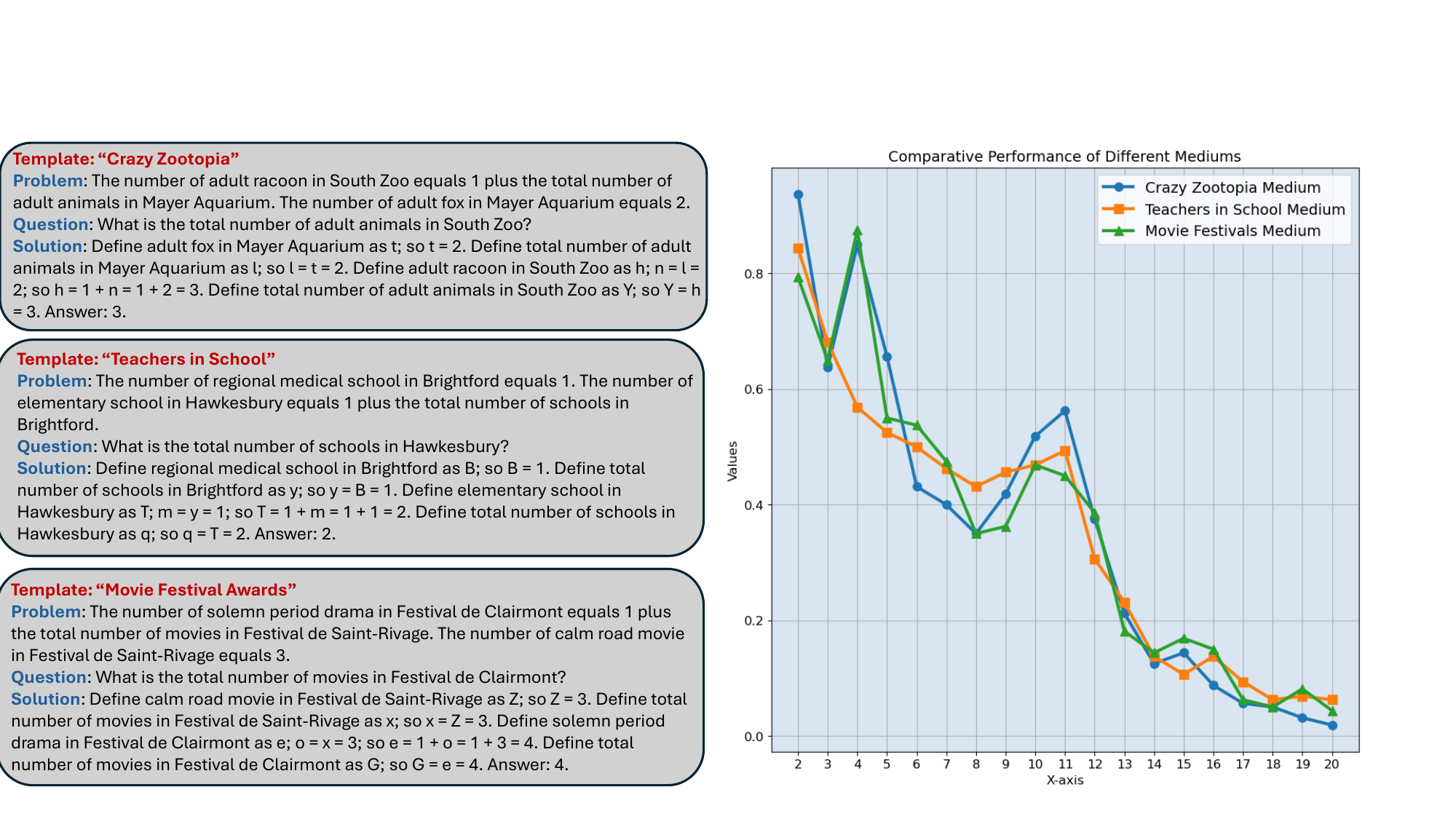} 
  \caption{Comparison between different task templates}
  \label{languagetemplates} 
\end{figure} 

We have three different real-world templates ready. We show that they offer consistent scores with Llama-3.1-8B-Instruct, with slight variables in specific operations. We also show three problem examples.

\newpage

\section{Long-Context Degradation of Models} 
\label{picutrall} 
In this section, we provide the accuracy decay curves of all LLMs tested across zero-context, 8K, 16K and 32K for further analysis. We selected the first 30 reasoning steps to truncate the data for comparison purposes. 

\begin{figure}[h!]
  \centering
  \begin{subfigure}{0.17\textwidth}
    \includegraphics[width=\textwidth]{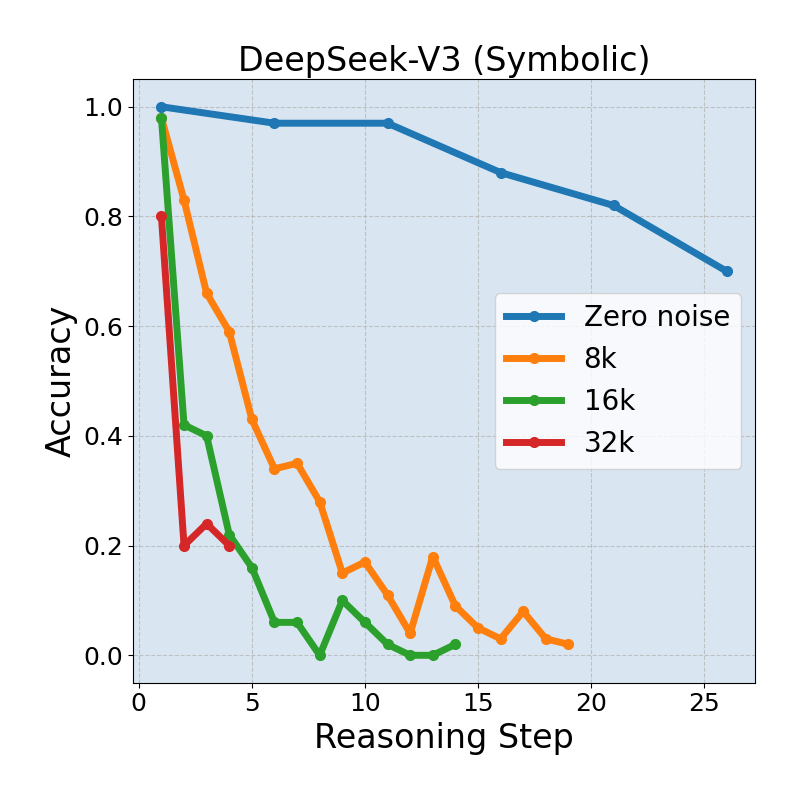}
  \end{subfigure}
  \hfill
  \begin{subfigure}{0.17\textwidth}
    \includegraphics[width=\textwidth]{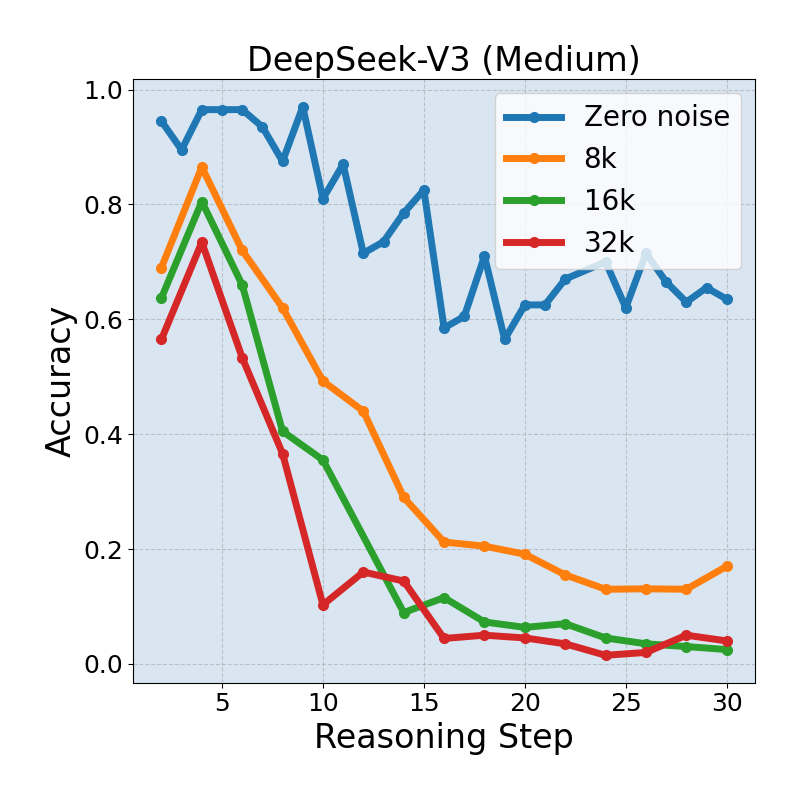}
  \end{subfigure}
  \hfill
  \begin{subfigure}{0.17\textwidth}
    \includegraphics[width=\textwidth]{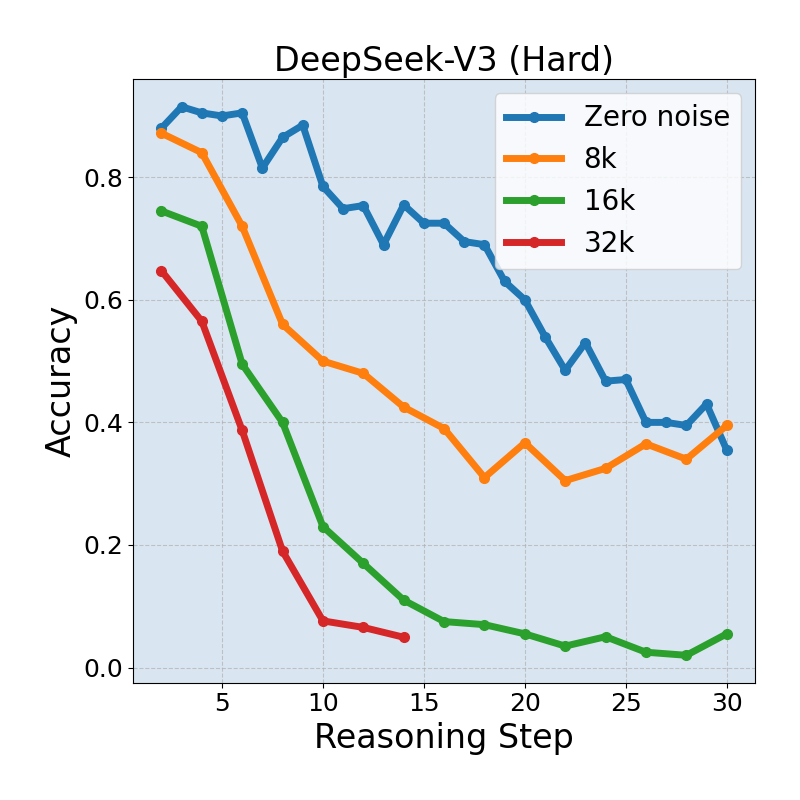}
  \end{subfigure}
  \hfill
  \begin{subfigure}{0.17\textwidth}
    \includegraphics[width=\textwidth]{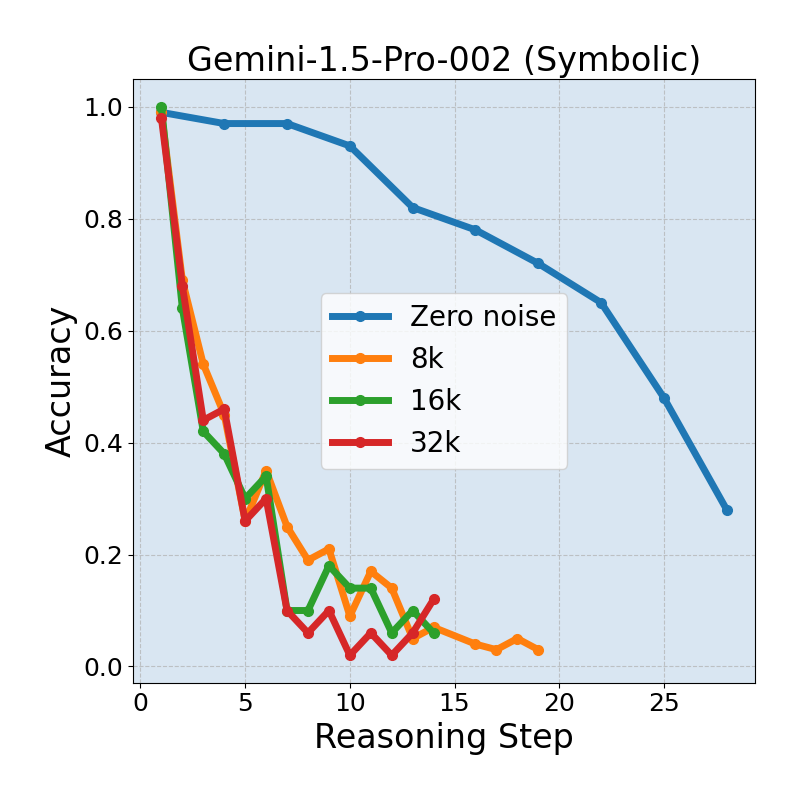}
  \end{subfigure}
  \hfill
  \begin{subfigure}{0.17\textwidth}
    \includegraphics[width=\textwidth]{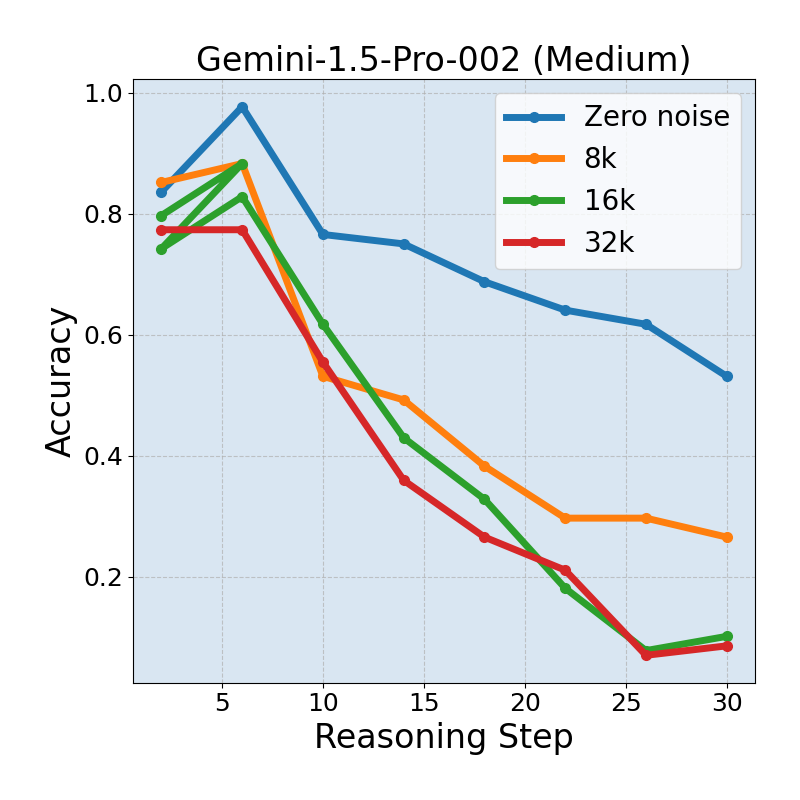}
  \end{subfigure}
  \hfill
  \begin{subfigure}{0.17\textwidth}
    \includegraphics[width=\textwidth]{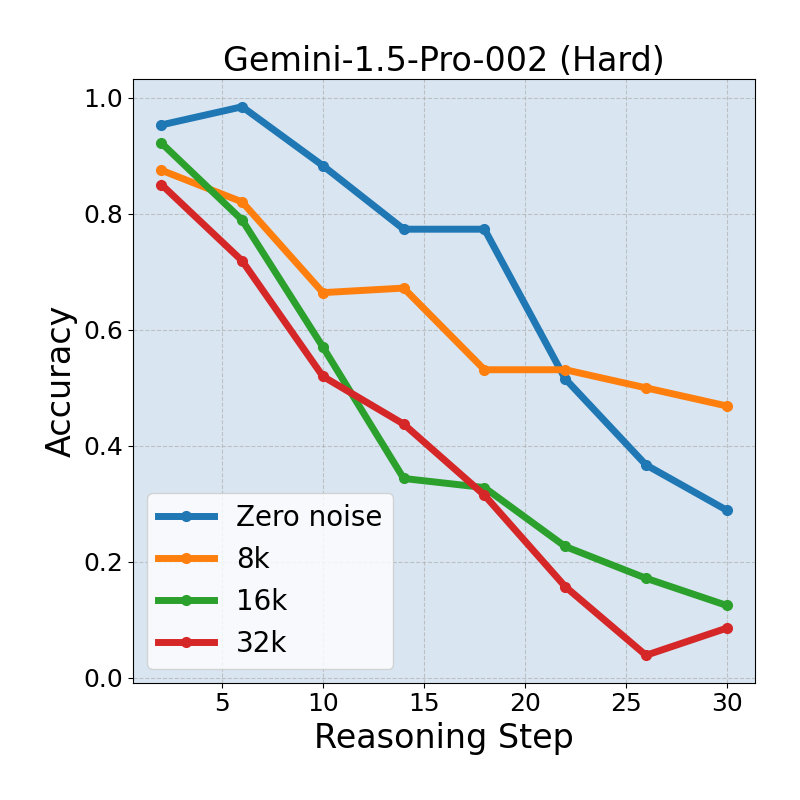}
  \end{subfigure}
  \hfill
  \begin{subfigure}{0.17\textwidth}
    \includegraphics[width=\textwidth]{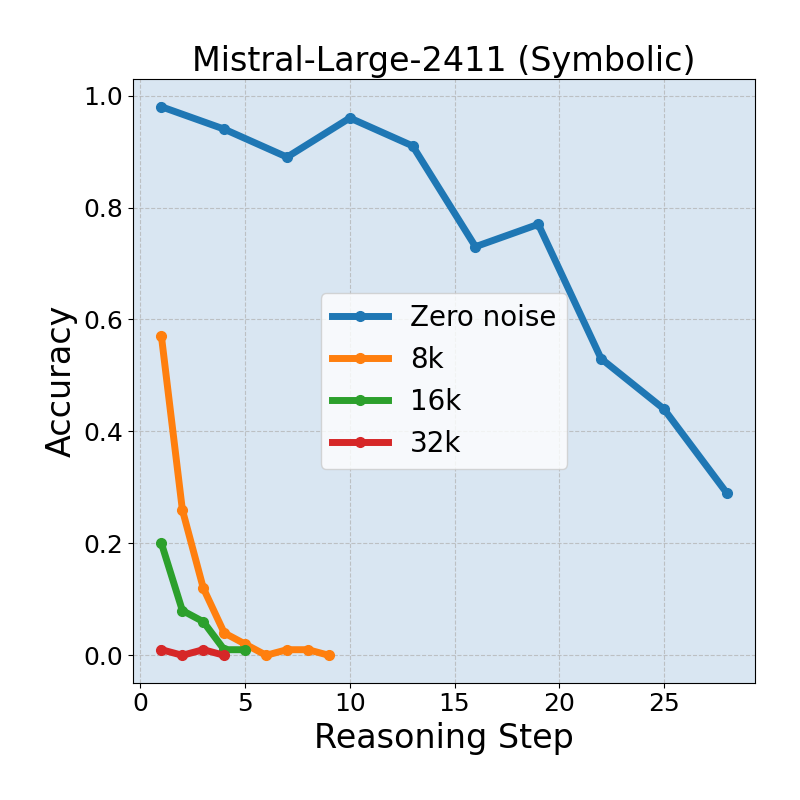}
  \end{subfigure}
  \hfill
  \begin{subfigure}{0.17\textwidth}
    \includegraphics[width=\textwidth]{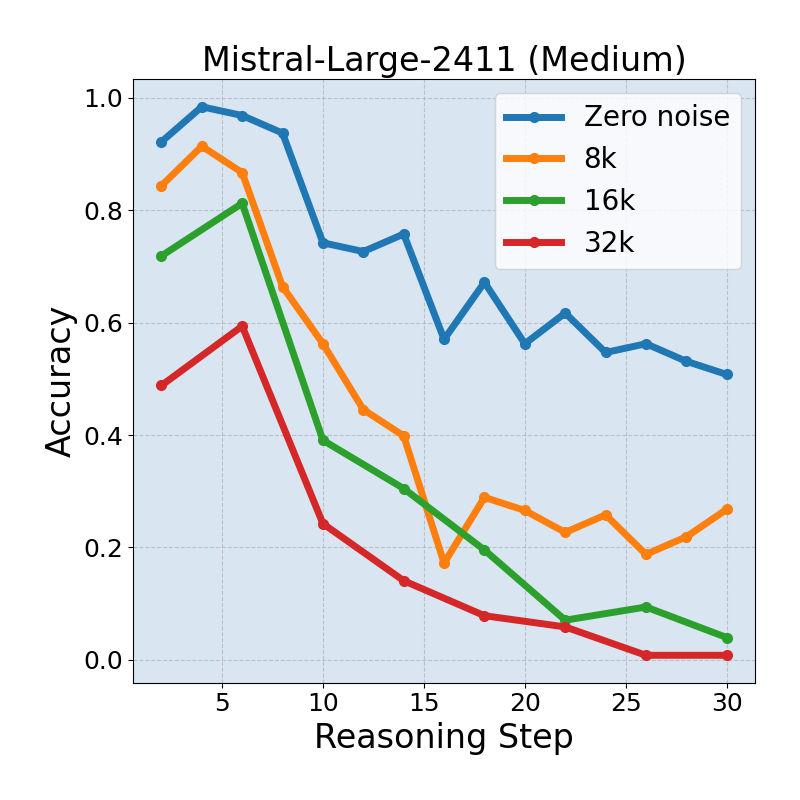}
  \end{subfigure}
  \hfill
  \begin{subfigure}{0.17\textwidth}
    \includegraphics[width=\textwidth]{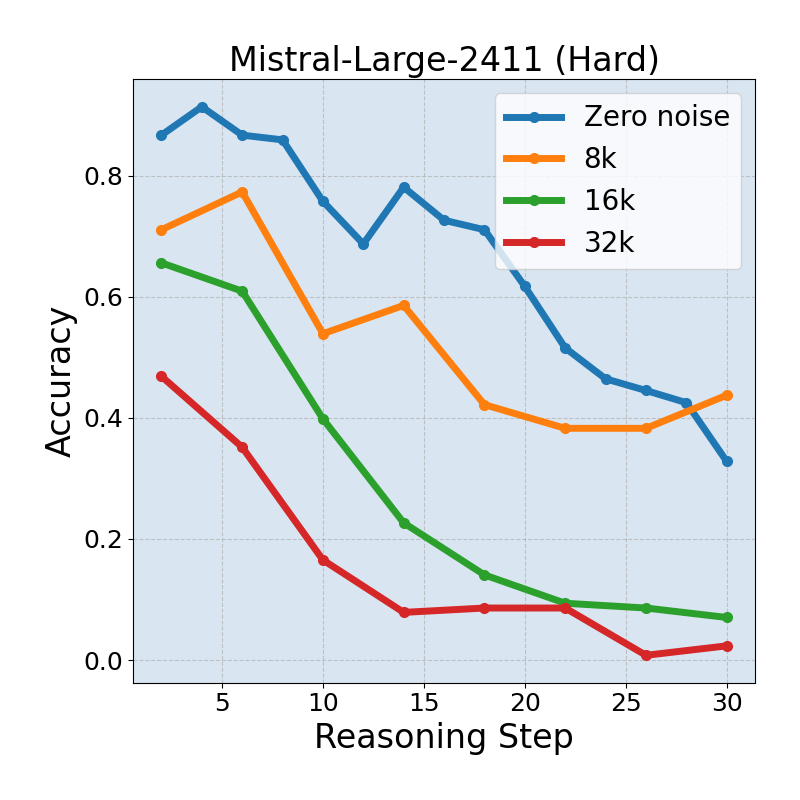}
  \end{subfigure}
  \hfill
  \begin{subfigure}{0.17\textwidth}
    \includegraphics[width=\textwidth]{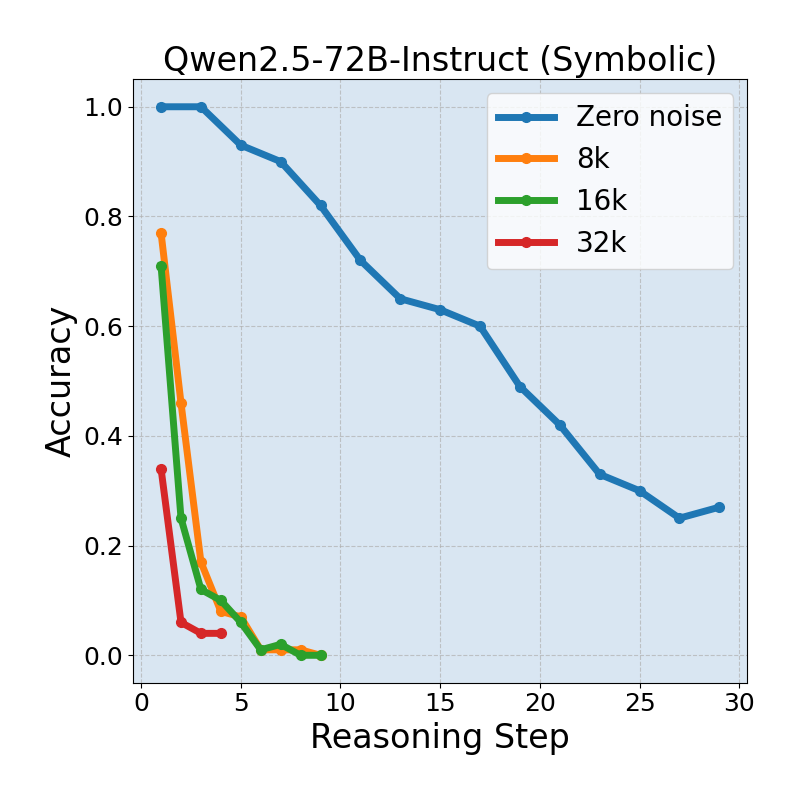}
  \end{subfigure}
  \hfill
  \begin{subfigure}{0.17\textwidth}
    \includegraphics[width=\textwidth]{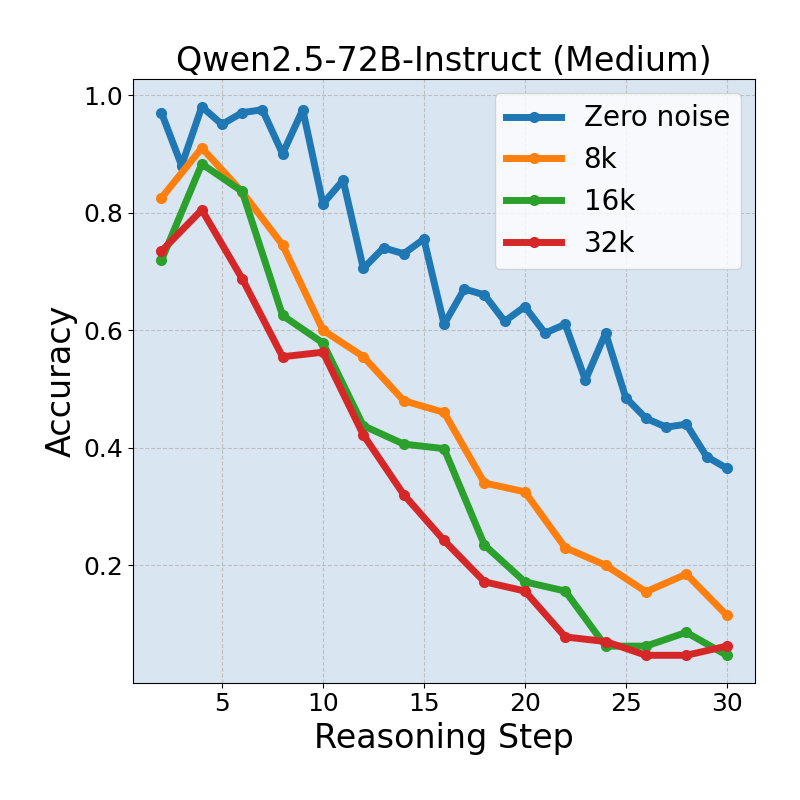}
  \end{subfigure}
  \hfill
  \begin{subfigure}{0.17\textwidth}
    \includegraphics[width=\textwidth]{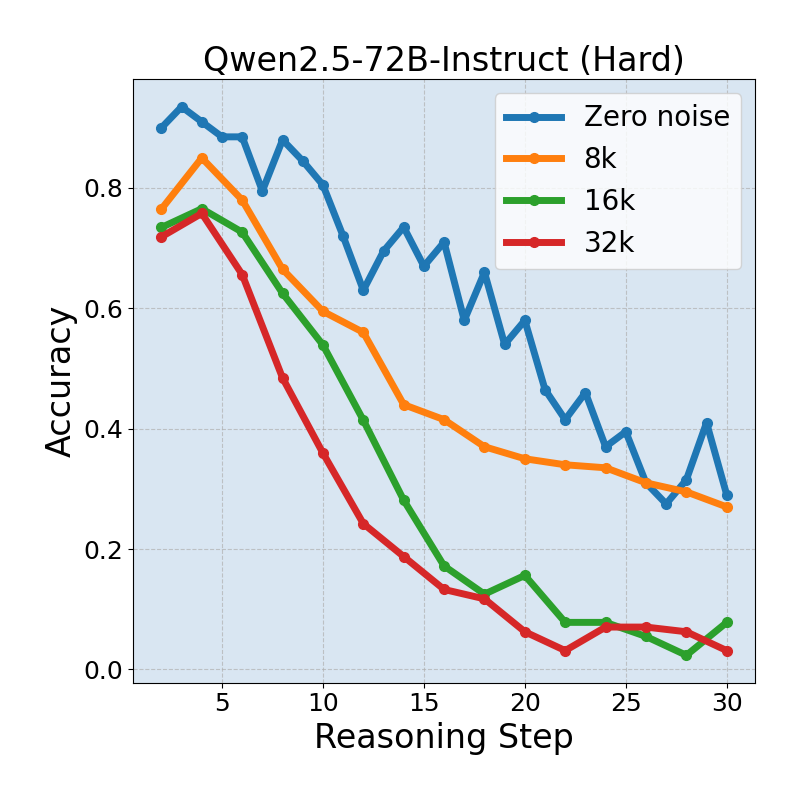}
  \end{subfigure}
  \hfill
  \begin{subfigure}{0.17\textwidth}
    \includegraphics[width=\textwidth]{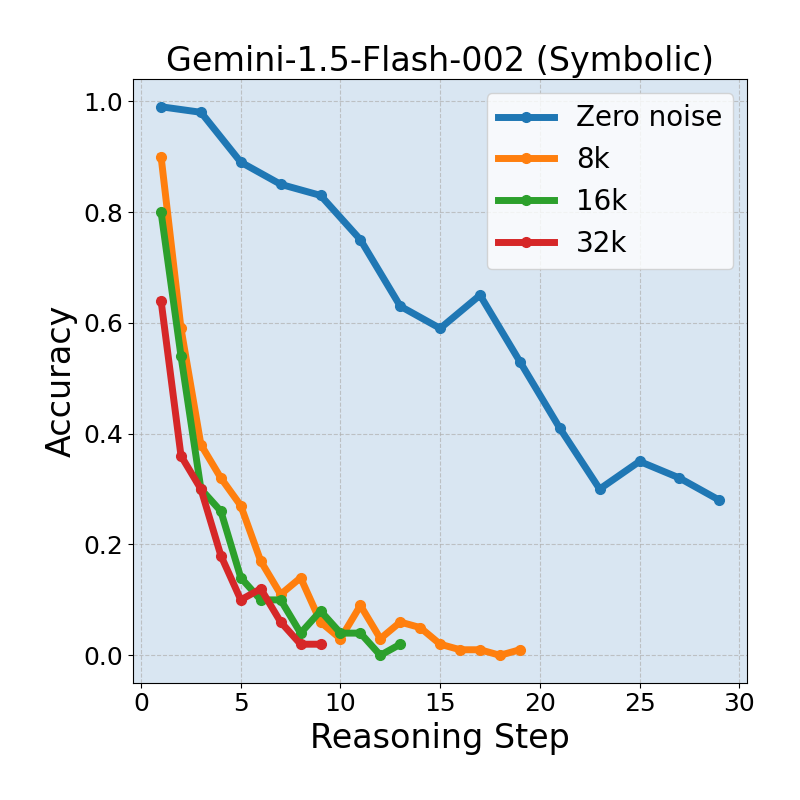}
  \end{subfigure}
  \hfill
  \begin{subfigure}{0.17\textwidth}
    \includegraphics[width=\textwidth]{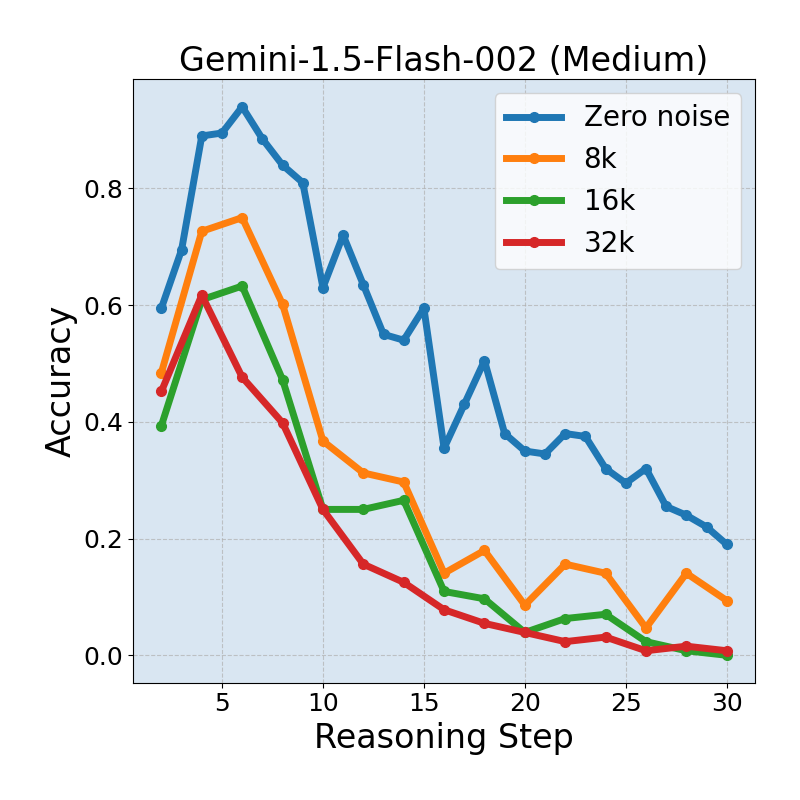}
  \end{subfigure}
  \hfill
  \begin{subfigure}{0.17\textwidth}
    \includegraphics[width=\textwidth]{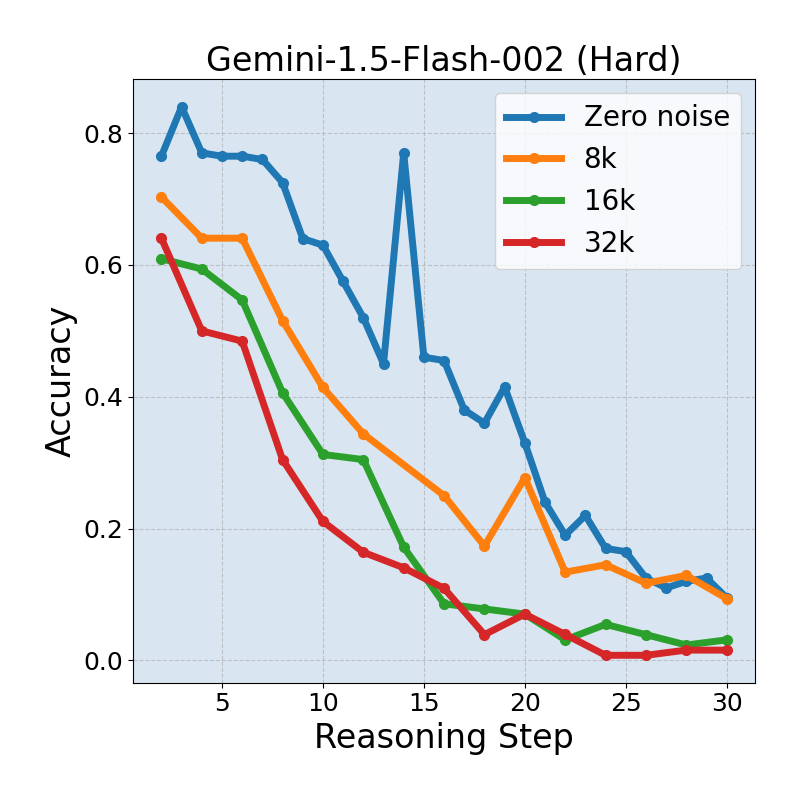}
  \end{subfigure}
  \hfill
  \begin{subfigure}{0.17\textwidth}
    \includegraphics[width=\textwidth]{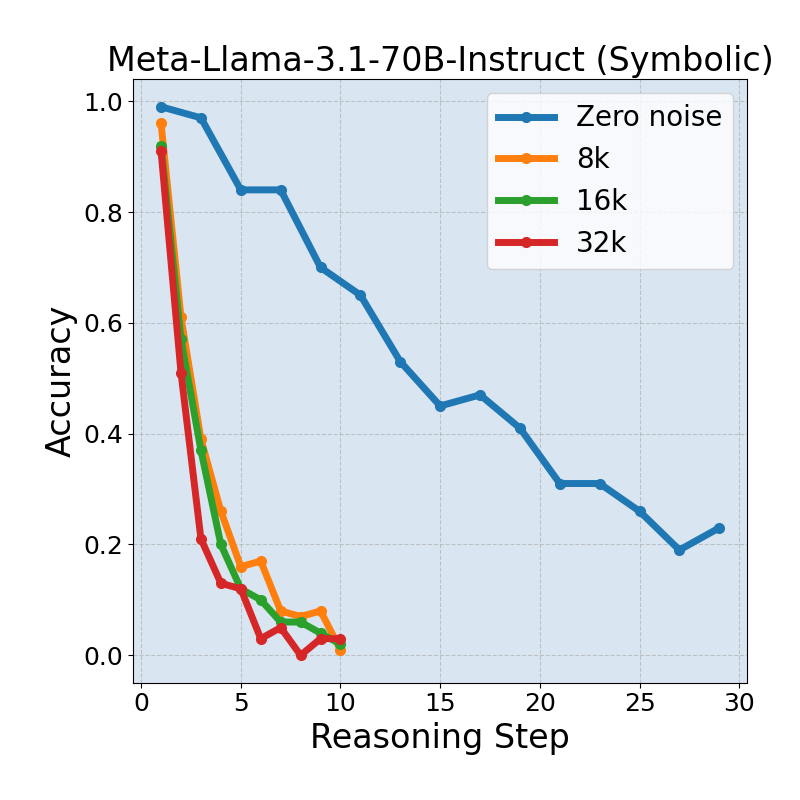}
  \end{subfigure}
  \hfill
  \begin{subfigure}{0.17\textwidth}
    \includegraphics[width=\textwidth]{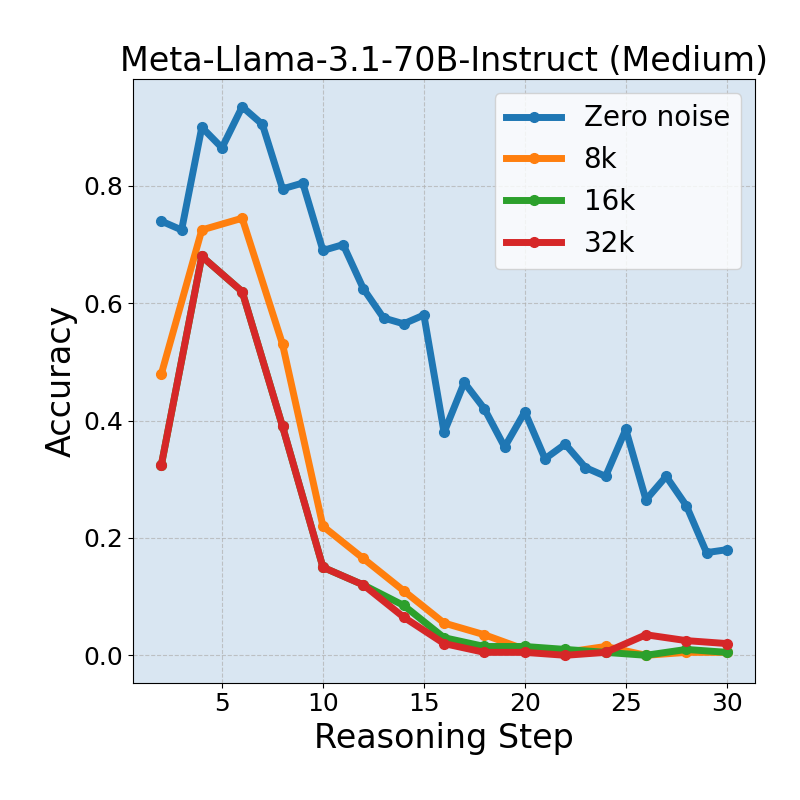}
  \end{subfigure}
  \hfill
  \begin{subfigure}{0.17\textwidth}
    \includegraphics[width=\textwidth]{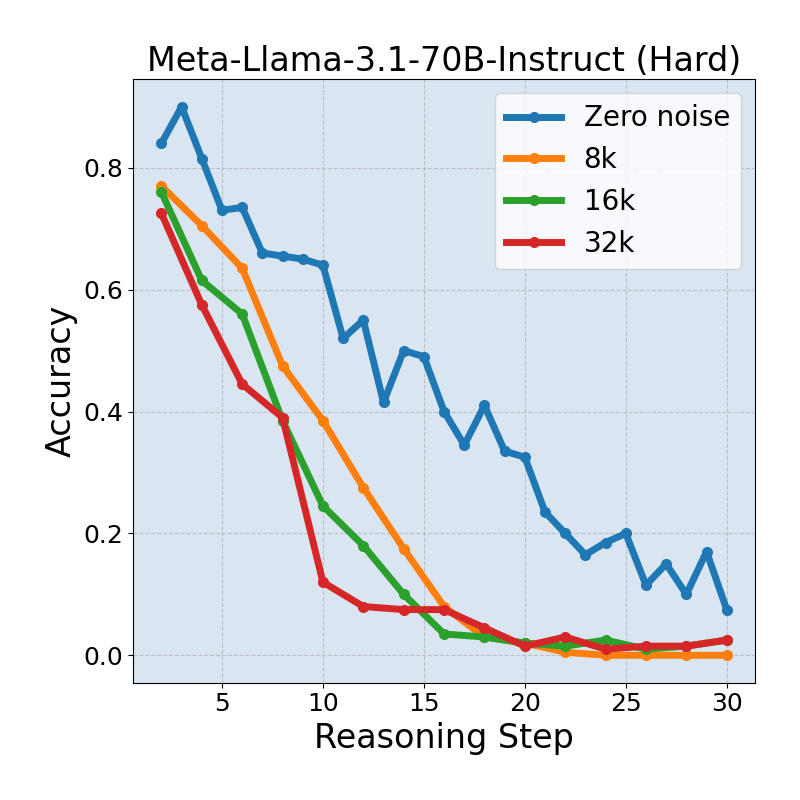}
  \end{subfigure}
  \hfill
  \begin{subfigure}{0.17\textwidth}
    \includegraphics[width=\textwidth]{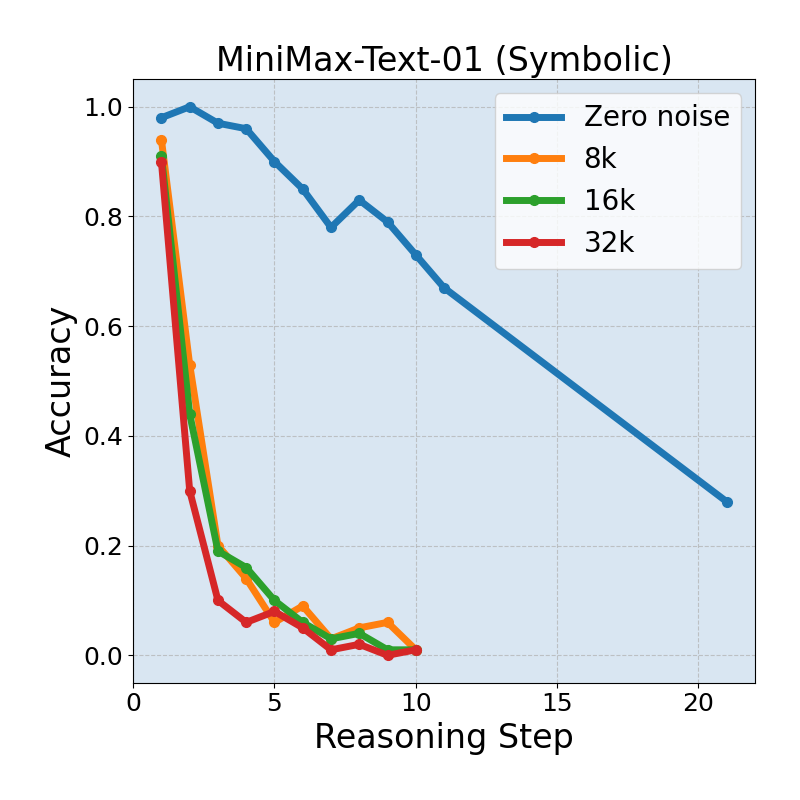}
  \end{subfigure}
  \hfill
  \begin{subfigure}{0.17\textwidth}
    \includegraphics[width=\textwidth]{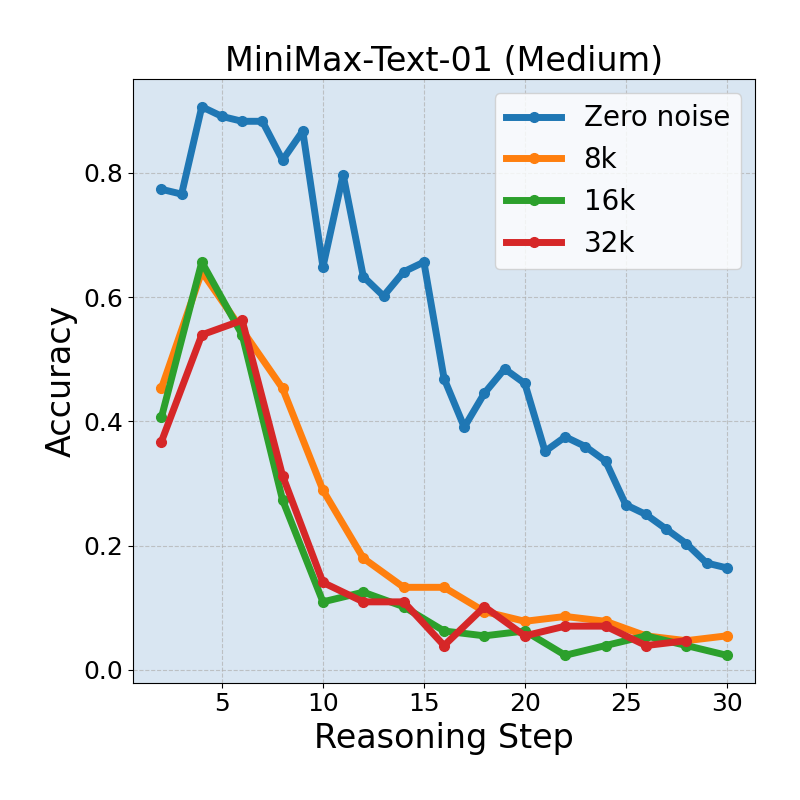}
  \end{subfigure}
  \hfill
  \begin{subfigure}{0.17\textwidth}
    \includegraphics[width=\textwidth]{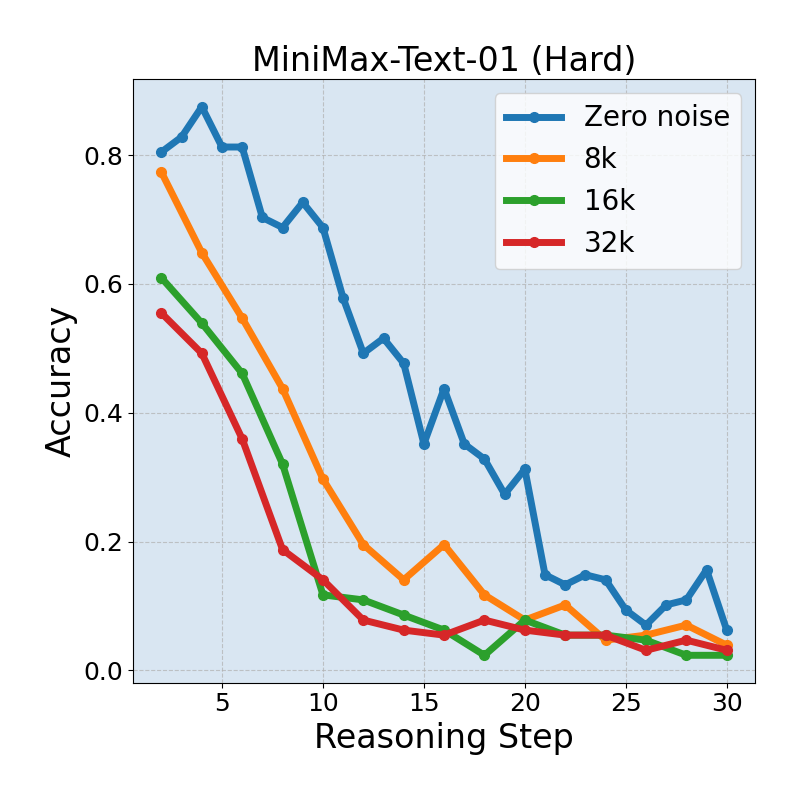}
  \end{subfigure}
  \hfill
  \begin{subfigure}{0.17\textwidth}
    \includegraphics[width=\textwidth]{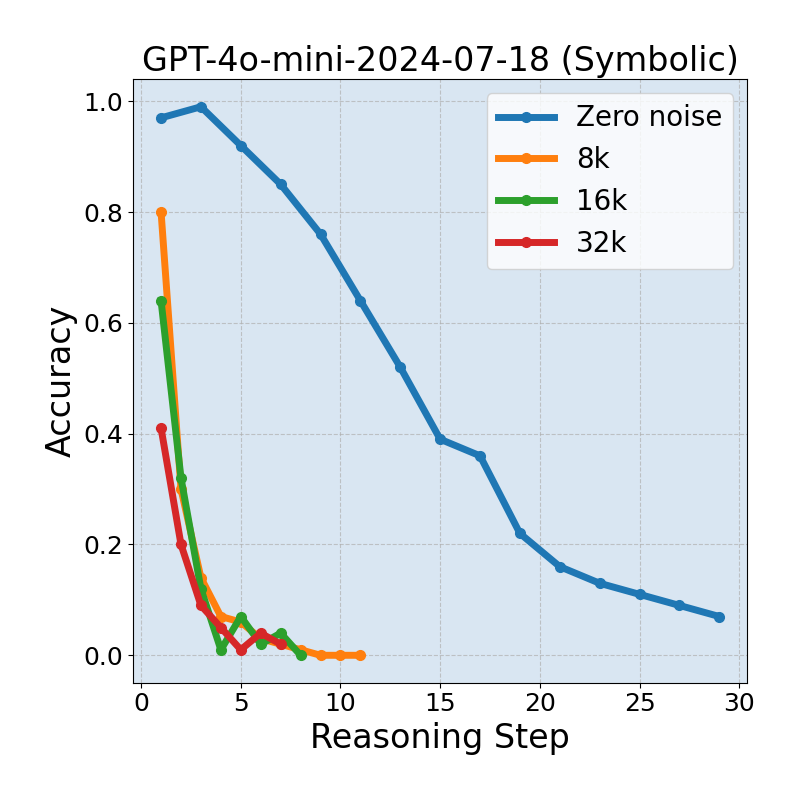}
  \end{subfigure}
  \hfill
  \begin{subfigure}{0.17\textwidth}
    \includegraphics[width=\textwidth]{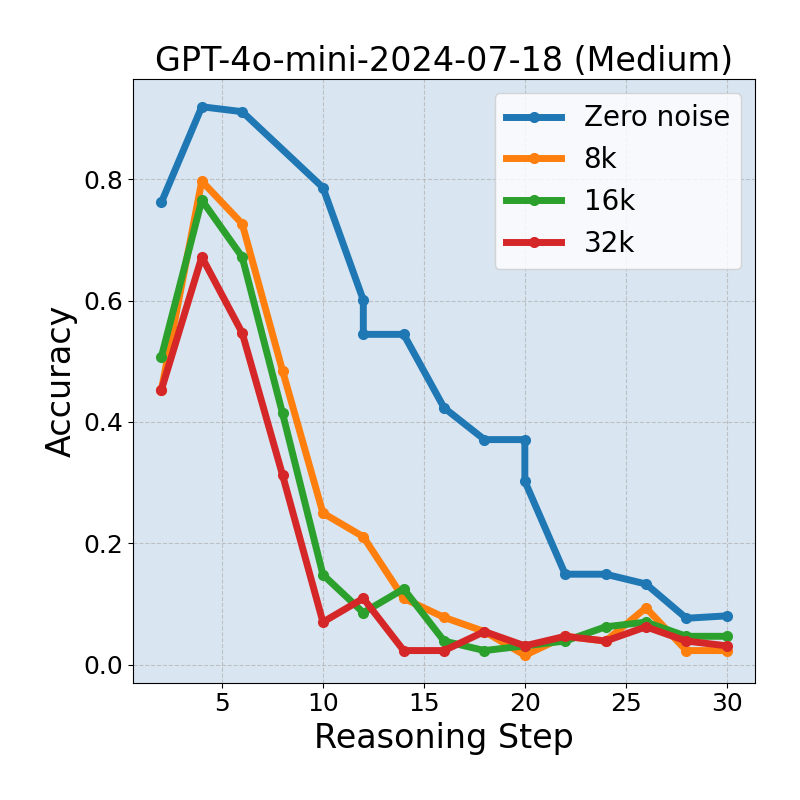}
  \end{subfigure}
  \hfill
  \begin{subfigure}{0.17\textwidth}
    \includegraphics[width=\textwidth]{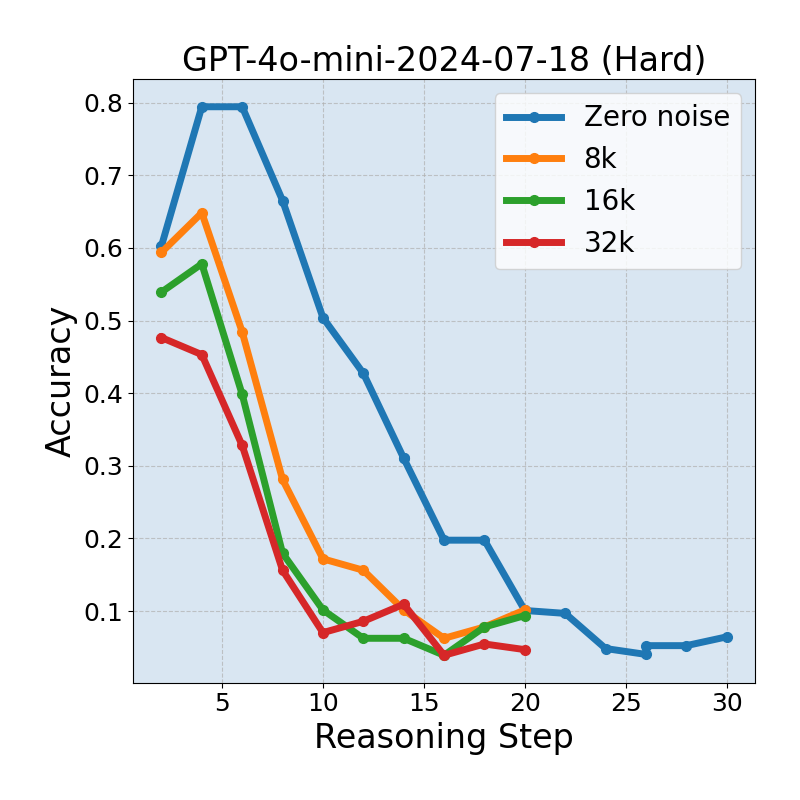}
  \end{subfigure}
  \hfill
  \begin{subfigure}{0.17\textwidth}
    \includegraphics[width=\textwidth]{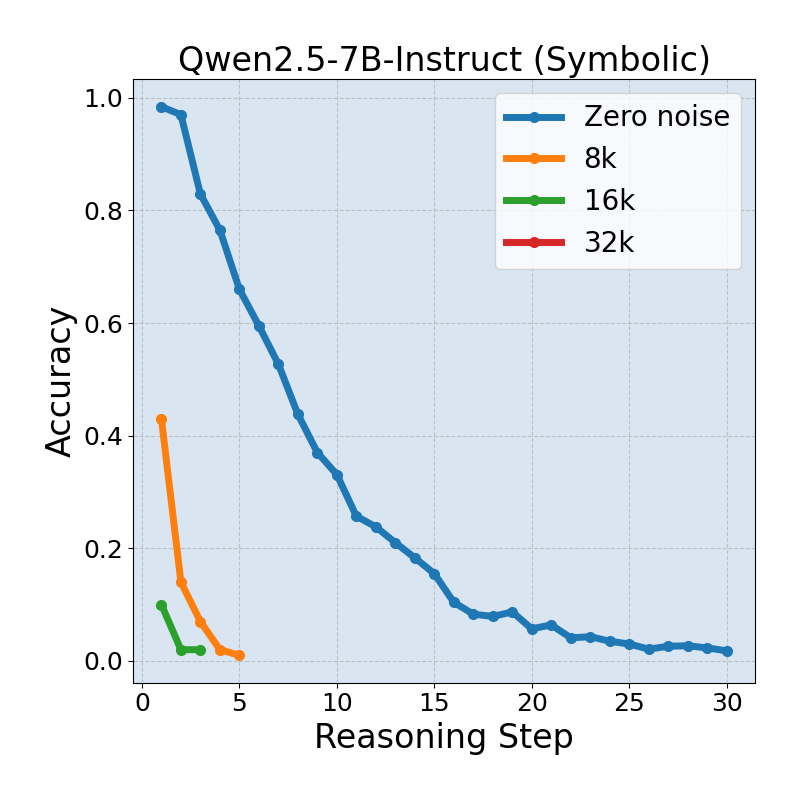}
  \end{subfigure}
  \hfill
  \begin{subfigure}{0.17\textwidth}
    \includegraphics[width=\textwidth]{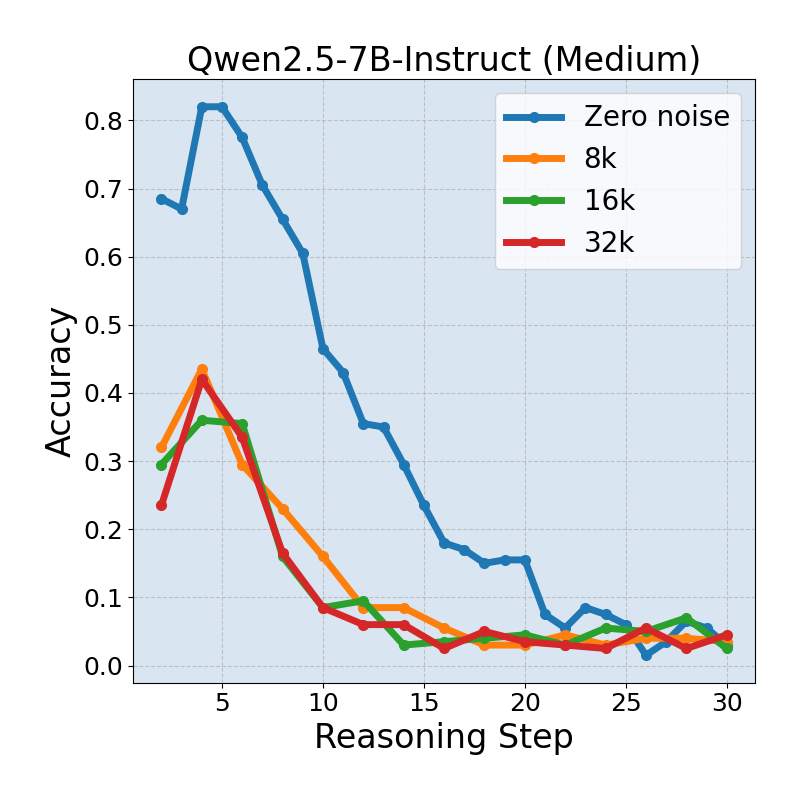}
  \end{subfigure}
  \hfill
  \begin{subfigure}{0.17\textwidth}
    \includegraphics[width=\textwidth]{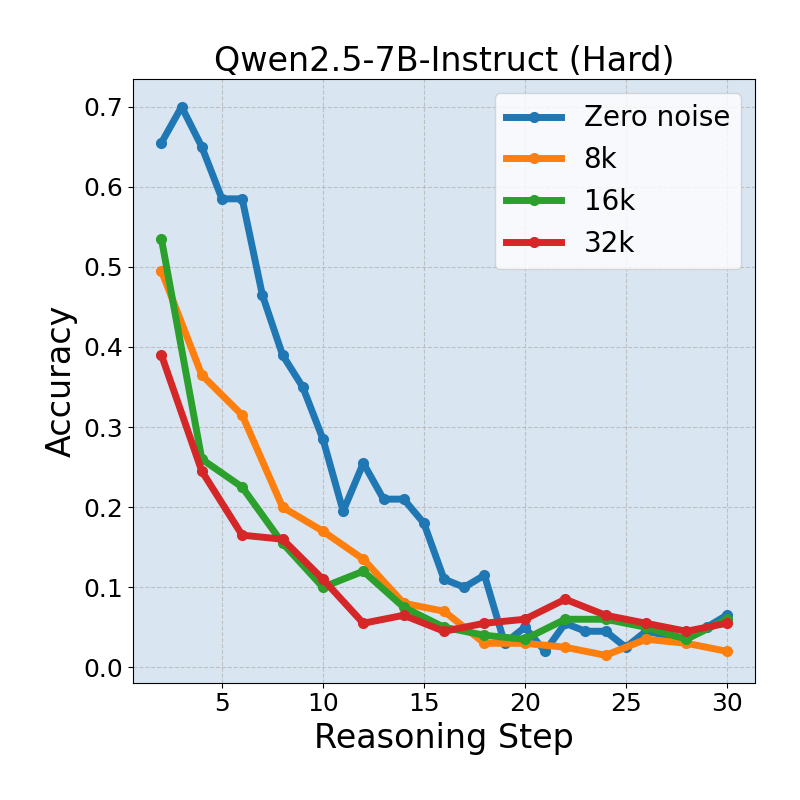}
  \end{subfigure}
  \hfill
  \begin{subfigure}{0.17\textwidth}
    \includegraphics[width=\textwidth]{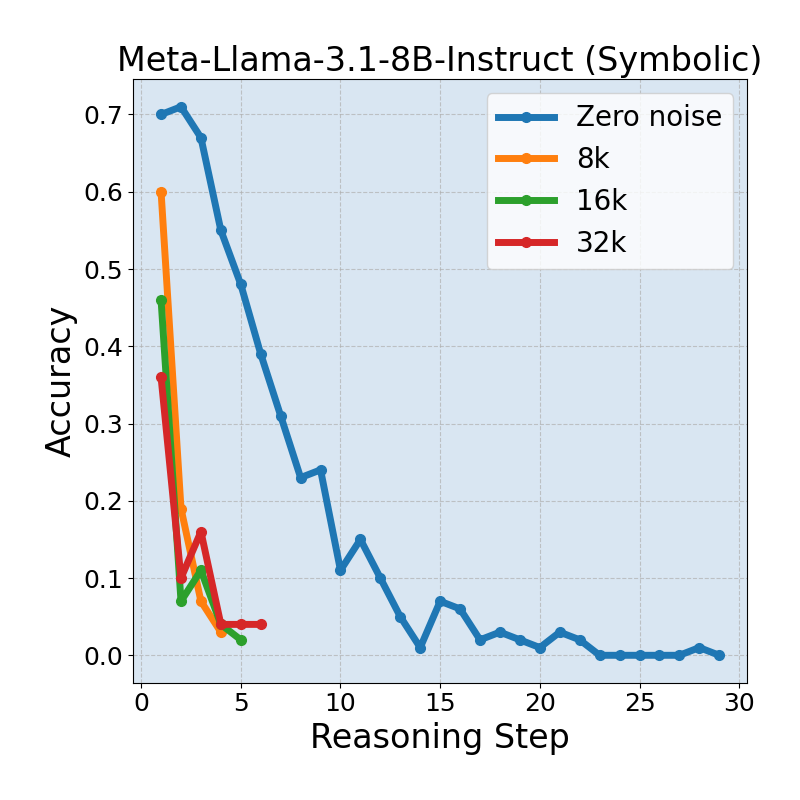}
  \end{subfigure}
  \hfill
  \begin{subfigure}{0.17\textwidth}
    \includegraphics[width=\textwidth]{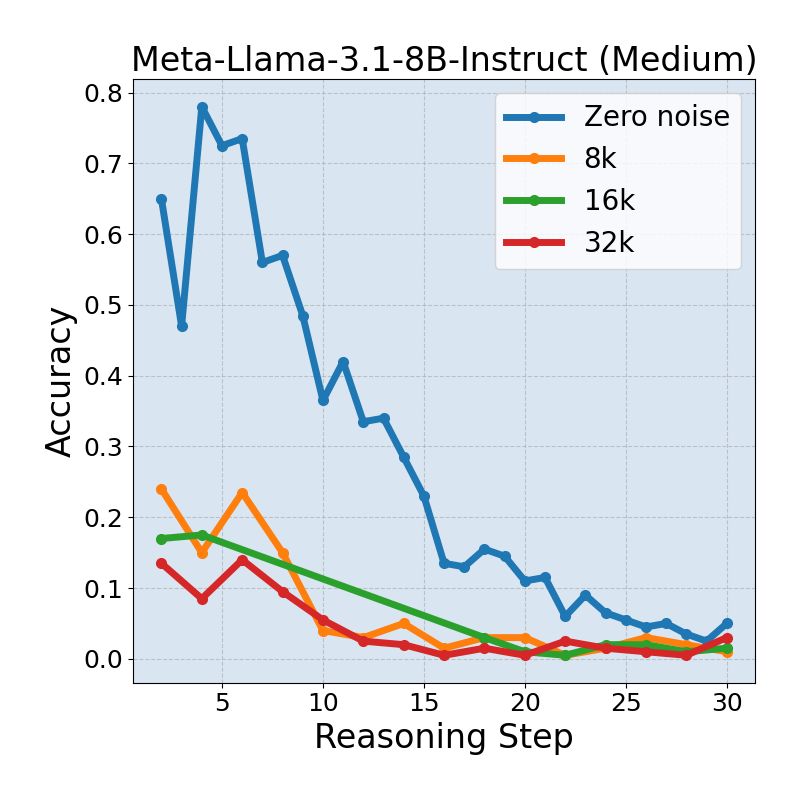}
  \end{subfigure}
  \hfill
  \begin{subfigure}{0.17\textwidth}
    \includegraphics[width=\textwidth]{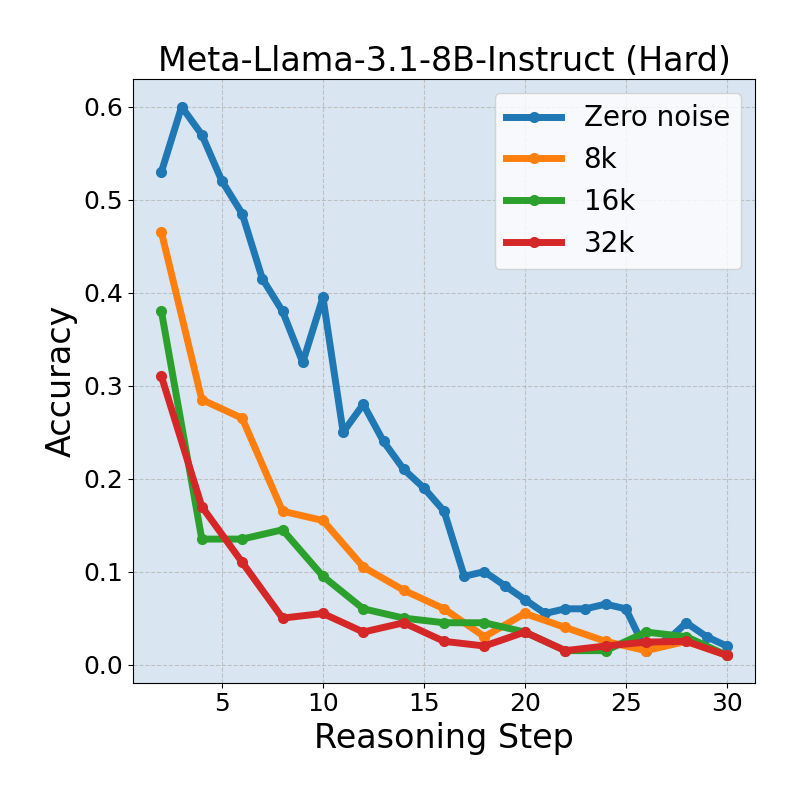}
  \end{subfigure}
  \caption{Accuracy decay with context length for different models}
\end{figure}

\end{document}